\documentclass[11pt]{article}
\usepackage[a4paper,margin=2.4cm]{geometry}
\usepackage[T1]{fontenc}
\usepackage{mathptmx,amsmath,amssymb,amsthm,graphicx,booktabs,xcolor,rotating}
\usepackage[numbers,sort&compress]{natbib}
\usepackage[colorlinks=true,linkcolor=blue!50!black,citecolor=blue!50!black,urlcolor=blue!50!black]{hyperref}
\newcommand{\pass}{\textcolor{green!45!black}{met}}
\newcommand{\fail}{\textcolor{red!70!black}{\textbf{not met}}}
\newtheorem{proposition}{Proposition}

\title{Predicting silent failures of learned inference\\ under restricted access: what measurement physics\\ and prior information explain}
\author{Mario Gianni\\ \small School of Computer Science and Informatics, University of Liverpool, UK\\ \small \texttt{Mario.Gianni@liverpool.ac.uk} \quad ORCID 0000-0001-5410-2377}
\date{18 October 2026}

\begin{document}
\maketitle

\begin{abstract}
Learned priors are used to reconstruct regions that measurements barely reach, and to report uncertainty there. They can fail silently: the reported interval excludes the truth while every ground-truth-free diagnostic stays within its noise. We call such errors the \emph{blind set} and ask whether they can be predicted before the target data are seen. For a correctly specified linear-Gaussian model averaged over the prior, the risk of a silent failure factorises exactly into non-detection and non-coverage, which does not hold for learned systems in general; we build a score that approximates this reference by linearising a VAE prior. In a prospective study on real X-ray CT objects (2DeteCT) under simulated restricted access, with predictions committed by hash before any trial, the score ranked the realised blindness of two learned inference systems (latent Laplace and ensemble Kalman inversion) with Spearman $\rho=0.79$ and $0.84$ over 240 pre-specified errors. Simple structural predictors would also have passed these tests. Sequential post-hoc analyses, each specified before its own computation, show that under bounded reach the learned score gave no detectable improvement over a physics-based baseline, in rankings or trial-level log scores. At the weakest attenuation studied, prior information improved rankings over the physics-based baseline, most clearly for ensemble Kalman inversion, but a Gaussian prior did as well and the gain was not detectable at the trial level. The score is not a calibrated probability. Silent failures can be anticipated here, but their structure follows mainly from measurement access and the decision quantity rather than from the learned prior.
\end{abstract}

\noindent\textbf{Keywords:} uncertainty quantification; silent failure; learned priors; variational autoencoder; Bayesian inverse problems; computed tomography; restricted access; pre-registration.

\section{Introduction}\label{sec:intro}

Reconstructing what cannot be observed directly, such as the interior of a sealed component or a region a sensor cannot reach, is increasingly done with learned priors \citep{bora2017}, and decisions are taken on the uncertainty these systems report. Such a system can fail in two ways. It can be uncertain and say so: its reported uncertainty is wide where its evidence is thin. Or it can be \emph{confidently wrong}: the reported uncertainty excludes the truth, while every check it can run without ground truth (residuals, consistency across measurement subsets, plausibility of the fitted latent state) stays within its noise. The second failure looks, from inside the system, exactly like a well-supported inference.

Each ingredient of this failure is established. Misspecification that the data cannot detect can be bounded or made harmless to estimators \citep{hansen2008,armstrong2021}; prior-data conflict checks find misspecification that does reach the data \citep{evans2006}. Priors mismatched to the truth can destroy the coverage of credible sets in inverse problems \citep{knapik2011,hadji2021}, including under partial observation \citep{gugushvili2018}. Learned reconstructions can place false structure in components the measurements do not determine \citep{bhadra2021}; learned uncertainty degrades under distribution shift \citep{ovadia2019}; conformal methods guarantee coverage only on average and under exchangeability \citep{angelopoulos2023}; and misspecified models can be precise yet inaccurate with little residual warning \citep{browning2026}. What is not established is whether the \emph{conjunction}---errors both undetected and outside the reported uncertainty---can be \emph{predicted in advance} for a learned non-linear system, from the model, the operator, the diagnostic suite and the noise alone. The blind set differs from each of these notions: non-coverage alone may be flagged by a diagnostic, an undetectable perturbation may leave the decision quantity unchanged, and prior-data conflict concerns misfit that the data do reveal. The blind set is their intersection, defined for a stated decision quantity and diagnostic suite.

We separate three claims: (a) blind errors can be predicted prospectively; (b) local linearisation of learned inference provides an effective predictor; (c) learned-prior geometry contributes information that simpler predictors lack. The evidence supports (a) in the tested setting, gives qualified support to (b), and does not establish (c) (Section~\ref{sec:disc}). The pre-specified tests show that a score computed before any trial ranks realised blindness. They do not separate it from structural baselines: the indicator of the unreached region and the error's effect on the decision quantity alone reach Spearman $\rho=0.52$--0.66 and 0.57--0.70 under bounded reach, so they too would have met the pre-specified threshold. Everything we conclude about \emph{what} makes blindness predictable comes from sequential post-hoc analyses, each specified in writing before its own computation but designed with earlier results known (Table~\ref{tab:levels}).

Linear-Gaussian pilots of the blind set are reported in a companion technical note \citep{gianni2026}; the present paper does not rely on them. Our contributions are of three kinds, which we keep apart.
\begin{enumerate}
\item \emph{Formulation and reference.} The blind set and its population-averaged risk (Section~\ref{sec:form}), and an exact result for the linear-Gaussian case (Proposition~\ref{prop:lg}): averaged over the prior, non-detection and non-coverage are independent for any detector, and the posterior spread is the correct scale. The result follows from a standard property of Gaussian conditioning; what it adds is its application to the blindness risk.
\item \emph{A computational procedure.} A prospective risk-ranking score that approximates this reference by linearising a learned prior (Section~\ref{sec:predict}). We offer no new theory of why learned uncertainty fails.
\item \emph{Empirical findings} about which information suffices to anticipate silent failure under different access regimes (Sections~\ref{sec:bounded}--\ref{sec:explain}): a prospective test on real CT objects with two learned inference systems; comparisons with physics-based, Gaussian and structural predictors at the level of rankings and of individual trials; and a full record of what was fixed when, of failed configurations and of deviations (Appendix~\ref{app:protocol}).
\end{enumerate}
Part of what we observe follows directly from the measurement null space, the size of an error and the sensitivity of the decision quantity; Proposition~\ref{prop:lg} makes this precise. The experiments add what does not follow from these alone: how often calibrated learned systems were silently wrong beyond reach; that, at the level of individual trials, the learned score gave no detectable predictive improvement over a physics-based score; that under weak but non-zero access prior information improves rankings, though not in a way specific to the learned prior nor as a detectable gain in trial-level predictive performance; and that a score computed before any target data, from a single surrogate, ranked the blindness of two different inference algorithms.

\section{Formulation and a linear-Gaussian reference}\label{sec:form}

\paragraph{Decision quantity, interval and detector.} The object is an image $x\in\mathbb R^{n}$ and the measurements are $y=Ax+\varepsilon$. The decision quantity is linear, $q(x)=c^\top x$, with $c=\mathbf 1_R/|R|$ the normalised indicator of a disc $R$, so $q$ is the mean attenuation over $R$. An inference system reports an estimate $\hat q(y)$, a spread $s(y)$ (the standard deviation it assigns to $q$) and a 95\% interval $I(y)=\hat q(y)\pm1.96\,s(y)$, where $1.96=\Phi^{-1}(0.975)$ is the 97.5\% quantile of the standard normal distribution, whose distribution function is $\Phi$. Ground-truth-free diagnostics are combined into a detector that raises an alarm when the data fall in a set $\mathcal A$ (Section~\ref{sec:design}). A trial is \emph{blind} when there is no alarm and $q(x)\notin I(y)$.

\paragraph{Blindness risk and blind set.} For a fixed object $x$ and a perturbation $\delta$ from an admissible class $\mathcal D$, the \emph{blindness risk} is
\[
\beta(\delta;x)=\Pr\big[y\notin\mathcal A\ \cap\ q(x+\delta)\notin I(y)\big],\qquad y=A(x+\delta)+\varepsilon,
\]
over the measurement noise and any randomness internal to the inference system. At level $\tau\in(0,1)$ the blind set is $B_\tau(x)=\{\delta\in\mathcal D:\beta(\delta;x)\geq\tau\}$. Since objects vary, the target used throughout is the \emph{population-averaged risk}
\[
\bar\beta_P(\delta)=\mathbb E_{x\sim P}\,\beta(\delta;x)
\]
for a stated population $P$. The score of Section~\ref{sec:predict} approximates $\bar\beta_P$ for $P$ the push-forward of the latent prior through the decoder; the experiments estimate $\bar\beta_P$ for $P$ the empirical distribution of a fixed set of held-out CT slices, with one or two noise draws per slice (the \emph{realised blind rate}). These two populations need not coincide (Section~\ref{sec:explain}), so our ranking results are empirical for the held-out population, not estimates of risk under other object distributions. The primary outcome is how well a score computed before any trial ranks errors by their realised blind rate, not membership of $B_\tau$.

\paragraph{An exact reference.} In a linear-Gaussian model the population-averaged risk has a closed form.

\begin{proposition}\label{prop:lg}
Let $x$ and $\varepsilon$ be independent Gaussian vectors, $y=Ax+\varepsilon$, and let $\hat q(y)=\mathbb E[q(x)\mid y]=a+L^\top y$ and $s^2=\operatorname{Var}[q(x)\mid y]$ (which does not depend on $y$). Let $\mathcal A$ be any measurable alarm set, and for a deterministic $\delta$ let $y_\delta=A(x+\delta)+\varepsilon$. Then, with probabilities over $x$ and $\varepsilon$,
\[
\Pr\big[y_\delta\notin\mathcal A\ \cap\ |q(x+\delta)-\hat q(y_\delta)|>1.96\,s\big]
=\Pr\big[y_\delta\notin\mathcal A\big]\,\Big\{\Phi\big(-1.96-\tfrac{e}{s}\big)+\Phi\big(-1.96+\tfrac{e}{s}\big)\Big\},
\]
where $e=c^\top\delta-L^\top A\delta$ is the error that $\delta$ induces in the estimate of $q$.
\end{proposition}

\begin{proof}
Write $y_0=Ax+\varepsilon$, so that $y_\delta=y_0+A\delta$ and $q(x+\delta)-\hat q(y_\delta)=\{q(x)-\hat q(y_0)\}+e$. In a jointly Gaussian model the posterior error $q(x)-\hat q(y_0)$ is independent of $y_0$ and distributed as $\mathcal N(0,s^2)$. The alarm event $\{y_\delta\notin\mathcal A\}=\{y_0\notin\mathcal A-A\delta\}$ is a function of $y_0$ alone. Hence the two events are independent, and the non-coverage probability is that of $\mathcal N(e,s^2)$ falling outside $\pm1.96\,s$.
\end{proof}

The proof rests on a standard property of Gaussian conditioning, that the posterior error is independent of the data; the proposition applies it to the blindness risk and is not a new probabilistic result. It holds for a correctly specified, jointly Gaussian model and for the risk averaged over the prior. It does not hold in general for non-linear learned inference or for a fixed object.

Four consequences frame the rest of the paper. (i) Averaged over the prior, the product of a non-detection and a non-coverage probability is exact, not an independence assumption, and the posterior spread $s$ (with no inflation) is the correct scale. (ii) If $A\delta=0$, the alarm probability equals the false-alarm rate whatever $\delta$ is, and the risk is a monotone function of $|c^\top\delta|/s$; within the null space, a ranking by effect size is therefore guaranteed by the model's assumptions. For any inference system, linear or not, an error with $A\delta=0$ leaves the data, and hence every diagnostic and the reported interval, unchanged (for a system with internal randomness, such as an ensemble method, it leaves their distribution unchanged); but whether the risk then increases with $|c^\top\delta|$ depends on the distribution of the system's own error, which the reference fixes as Gaussian and independent of the data. For the realised non-linear systems it is not guaranteed. (iii) For a \emph{fixed} object neither statement holds in general: the estimate is shrunk towards the prior mean, so the error is offset from $e$ by a deterministic shrinkage bias, its sampling variance is smaller than $s^2$ (Section~\ref{sec:predict}), and it is correlated with the residual. (iv) If the detector is a Mahalanobis test on diagnostics $T$ that are linear in $y$ with null covariance $\Sigma_T$, a mean shift $\Delta$ gives an alarm probability $\Pr[\chi^2_\nu(\Delta^\top\Sigma_T^{-1}\Delta)>t]$ exactly; using one standardised component instead can only understate the noncentrality, since $\Delta^\top\Sigma_T^{-1}\Delta\geq\Delta_1^2/(\Sigma_T)_{11}$. Our detector includes diagnostics that are quadratic in $y$, so this law is itself an approximation for it. A Monte Carlo check under the linearised model reproduces the proposition numerically, which verifies the implementation (Section~\ref{sec:explain}).

\section{The prospective score}\label{sec:predict}

The score uses only the learned prior, the operator, the diagnostics and the noise model. A decoder $G:\mathbb R^{32}\to\mathbb R^n$ with latent prior $\mathcal N(0,I)$ defines the learned prior (Section~\ref{sec:design}). We draw $K=100$ linearisation points $z_k\sim\mathcal N(0,I)$ and linearise the VAE-Laplace system at each: with $A_w=L_y^{-1}A$ ($L_y$ the Cholesky factor of the data covariance $\Sigma_y$, which includes decoder model error), $J_k=A_w\,\partial G/\partial z(z_k)$, $\Sigma_{z,k}=(J_k^\top J_k+I)^{-1}$ and gain $K_k=\Sigma_{z,k}J_k^\top$. This is the local linear-Gaussian model of Proposition~\ref{prop:lg}. For a pixel-space error $\delta$ added to the truth, the linearised response is a residual shift $r_k=(I-J_kK_k)A_w\delta$, with detectability $d_k=\|r_k\|^2$, and an error in the estimate of $q$,
\[
e_k=\big(c-A_w^\top K_k^\top g_k\big)^\top\delta,\qquad g_k=\partial G/\partial z(z_k)^\top c,
\]
to be compared with the spread $s_k=\kappa\,(g_k^\top\Sigma_{z,k}g_k+v_q)^{1/2}$, where $v_q$ is the variance of the decoder model error in $q$ and $\kappa$ is the system's interval inflation (Section~\ref{sec:design}). The local probabilities are
\[
P_{\mathrm{uncov},k}=\Phi(-1.96-e_k/s_k)+\Phi(-1.96+e_k/s_k),\qquad
P_{\mathrm{det},k}=\Pr\big[\chi^2_{\nu}(\lambda_k)>t\big],
\]
with $t$ the realised detector's empirical threshold and $\nu=9$, the number of retained diagnostics. The detector's null covariance has rank 8, because $T_0$ is a linear combination of the six block residuals, so the rank, not 9, is the appropriate dimension; the frozen score keeps $\nu=9$, and the consequence is examined in Section~\ref{sec:explain}. The noncentrality uses the total-residual diagnostic $T_0=(\|r\|^2-m)/\sqrt{2m}$, with $m$ the number of rays: an error shifts the mean of $\|r\|^2$ by $d_k$, so the mean of $T_0$ by $d_k/\sqrt{2m}$, and dividing by the null standard deviation $s_0$ of $T_0$ gives a standardised shift whose square is $\lambda_k=\{d_k/(\sqrt{2m}\,s_0)\}^2$. The shifts of the other diagnostics are ignored. The \emph{blindness risk score} is
\[
S_{\mathrm{blind}}(\delta)=\frac1K\sum_{k=1}^K P_{\mathrm{uncov},k}\,(1-P_{\mathrm{det},k}),
\]
a number in $[0,1]$ used to rank errors. The analysis plans and deposited files call it $P_{\mathrm{blind}}$; we rename it because it is not a calibrated probability (Section~\ref{sec:explain}). No target data enter: for a transplanted object, $\delta$ is the patch minus the mean training image under its mask, since the base it will meet is unknown. The same score, computed with the Laplace linearisation and the Laplace $\kappa$, is used for \emph{both} learned systems; for EKI it is a transferred surrogate, not a system-specific predictor.

\paragraph{Departures from the reference.} With $\kappa=1$, the non-coverage term and the product form are exact for the prior-averaged risk of the local model (Proposition~\ref{prop:lg}); the detection term would be exact for a Mahalanobis test on diagnostics linear in the data if it used the full noncentrality (consequence (iv)). The score departs from this reference in five ways, which the experiments probe. (1) \emph{Non-linearity}: the average over $z_k$ combines local models across prior draws of the operating point, so $S_{\mathrm{blind}}$ is a prior-averaged local sensitivity score, not a posterior predictive probability. (2) \emph{Error scale}: $s_k$ includes $\kappa$, chosen for empirical marginal coverage (Section~\ref{sec:design}); it does not show that each local spread describes the error at that point, nor that the error is Gaussian. (3) \emph{Detection}: the one-component noncentrality is a heuristic for a detector with partly quadratic diagnostics. (4) \emph{Population}: the experiments use held-out CT slices, not draws from $G(z)$, $z\sim\mathcal N(0,I)$. (5) \emph{Decoder misspecification}, modelled only through a stationary model-error covariance.

\paragraph{Fixed objects.} Take an object whose latent state is the linearisation point $z_k$. Under the prior $\mathcal N(0,I)$ the estimate is shrunk towards zero: in whitened form $\hat z-z_k=-\Sigma_{z,k}z_k+K_k\varepsilon$. The error in $q$ is therefore $q-\hat q=e_k+g_k^\top\Sigma_{z,k}z_k-g_k^\top K_k\varepsilon-\xi$, with $\varepsilon$ whitened noise and $\xi$ the model error in $q$: the induced error $e_k$ plus a deterministic shrinkage bias. Its sampling variance is $g_k^\top K_kK_k^\top g_k+v_q$; the posterior variance is $g_k^\top\Sigma_{z,k}g_k+v_q$. Because $\Sigma_{z,k}=\Sigma_{z,k}(J_k^\top J_k+I)\Sigma_{z,k}$ and $K_kK_k^\top=\Sigma_{z,k}J_k^\top J_k\Sigma_{z,k}$,
\[
\Sigma_{z,k}=K_kK_k^\top+\Sigma_{z,k}^2 ,
\]
so the posterior variance exceeds the sampling variance by exactly $g_k^\top\Sigma_{z,k}^2g_k\geq0$, which for $z_k\sim\mathcal N(0,I)$ is the variance of the shrinkage bias. The posterior variance is thus the sampling variance plus the prior-averaged variance of the bias. For a fixed object the error is not centred at $e_k$, and the offset is largest where $\Sigma_{z,k}$ is large. Where the data dominate $q$ the two nearly coincide; where the prior sets $q$, as beyond reach, they differ, and only the prior-averaged statement of Proposition~\ref{prop:lg} justifies the posterior scale.

\paragraph{Baselines.} The \emph{Gaussian prediction} is the same construction applied to the Gaussian baseline system (Section~\ref{sec:design}): its own linear blindness model, with its own $\kappa$ and detector, used to rank the learned systems' blindness. The \emph{physics-based baseline}, $P_{\mathrm{uncov}}(|c^\top\delta|/\bar s)\,\{1-P_{\mathrm{det}}(\|A_w\delta\|^2)\}$, uses the operator, the noise and model-error covariance in $A_w$ and the decision functional, with a fixed scale $\bar s$ (the mean VAE-Laplace spread), but no learned-prior geometry and no learned fit; its whitening and scale are derived from the VAE. A \emph{noise-only physics baseline} has neither (noise-only whitening, a simple residual test and the prior standard deviation $\sigma_q=0.041$ of $q$). The full-diagnostic variant \emph{v2} propagates the mean shift of all nine diagnostics through the realised detector's null; it is reported as robustness, and the pre-specified score remains v1. Finally, \emph{structural predictors} are the effect size $|c^\top\delta|$ and the indicator of the unreached region.

\section{Experimental design}\label{sec:design}

\paragraph{Objects.} We use the high-fidelity (mode 2) reference reconstructions of 2DeteCT \citep{kiss2023}, slices 1--2000, block-mean downsampled to $64\times64$ and multiplied by 100. Consecutive slices image the same sample tube and are strongly correlated, so a contiguous split is not exchangeable (Appendix~\ref{app:protocol}). We split by interleaved 10-slice blocks (index $\equiv4\pmod{10}$ for calibration, $\equiv9$ for testing, the rest for training), removing from training the two slices either side of every held-out block: 1,442 training, 200 calibration and 200 test slices. Every sample tube therefore contributes to training and to testing. The 750 out-of-distribution slices of 2DeteCT (figs, nuts, titanium screws and other objects absent from the training mixtures) are used only to build errors.

\begin{figure}[t]\centering
\includegraphics[width=\linewidth]{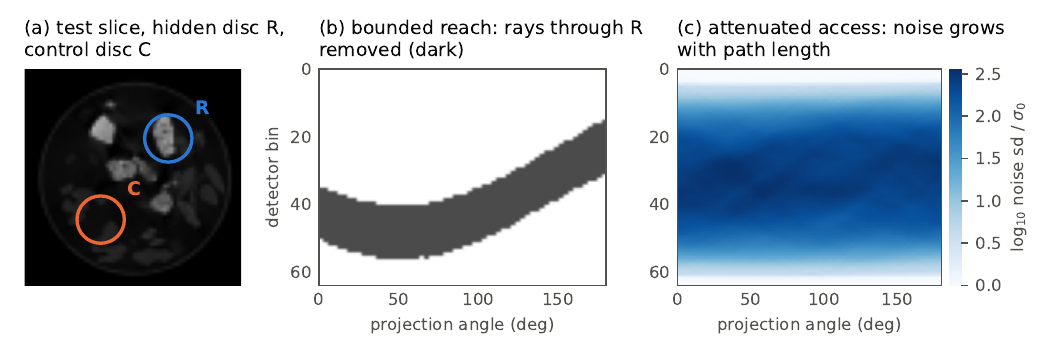}
\caption{Set-up. (a) A test slice with the hidden disc $R$ and the within-reach control disc $C$. (b) Bounded reach: every ray through $R$ is removed (dark band in the sinogram). (c) Attenuated access (study A3): no ray is removed, but the noise standard deviation grows with path length, up to $\approx$360\,$\sigma_0$ on the longest paths ($F=128{,}000$).}\label{fig:setup}
\end{figure}

\paragraph{Access.} Measurements are simulated with a parallel-beam CT operator (60 angles over $180^\circ$, 64 bins; 3,840 rays) with Gaussian noise of standard deviation $\sigma_0$ equal to 1\% of the RMS projection. Under \emph{bounded reach}, every ray through a hidden disc $R$ (radius 7~px) is removed, so $R$ lies exactly in the null space of the operator (Fig.~\ref{fig:setup}a,b; $m=2{,}891$ rays kept, 2,896 in the second geometry). Under \emph{attenuated access} no ray is removed ($m=3{,}840$); the noise variance grows with path length, $\sigma_i^2=\sigma_0^2\exp(\bar p_i/\ell)$, with $\bar p_i$ the projection of the mean training image along ray $i$ and length scale $\ell=\max_i\bar p_i/\ln F$. Thus $\sigma_i^2=\sigma_0^2F^{\bar p_i/\max\bar p}$, and the attenuation strength $F$ is the noise-variance multiplier on the longest path (Fig.~\ref{fig:setup}c). This Gaussian model is a simplification of photon-starved CT, whose counts are Poisson before the log transform.

\paragraph{Decision quantity.} Under bounded reach $c$ is the normalised indicator of $R$; under attenuated access, of a central disc of the same radius. A control quantity over a within-reach disc $C$ checks calibration where the data are informative.

\paragraph{Learned prior.} A convolutional VAE \citep{kingma2014} with latent dimension 32 (encoder: four stride-2 $4\times4$ convolutions with 32, 64, 128, 128 channels and a 256-unit dense layer; decoder: dense layers to $4\times4\times128$, four nearest-neighbour upsampling and $3\times3$ convolution stages with 128, 64, 32, 16 channels; GELU) is trained on the training slices with the eight symmetries of the square (2,000 Adam steps, batch 64, cosine-decayed learning rate from $10^{-3}$, decoder noise 0.05). The KL weight rose linearly from 0 to 1 over the whole run, reaching 1 only at the last step, so the encoder's aggregate posterior need not match $\mathcal N(0,I)$. The decoder is imperfect, and we model its error as a stationary Gaussian field whose covariance is the radially averaged power spectrum of the decoder residuals on the calibration slices (16 parameters), giving $\Sigma_y=A\,C_{\mathrm{pix}}A^\top+\mathrm{diag}(\sigma_i^2)$ and the bias (0.0004) and standard deviation (0.017) of the model error in $q$. Stationarity is an assumption, since decoder errors concentrate near object edges; it was chosen because a rank-60 empirical covariance over-fitted the detector null (Appendix~\ref{app:protocol}).

\paragraph{Inference systems.} (i) \emph{VAE-Laplace}: latent MAP by damped Gauss--Newton (up to 30 iterations) with a Laplace approximation in latent space. It is warm-started from the encoder applied to the Gaussian-baseline posterior mean, which couples the two systems; a cold start was not tested. (ii) \emph{VAE-EKI}: ensemble Kalman inversion in latent space \citep{iglesias2013}, ES-MDA form \citep{emerick2013}, 200 members, four steps with inflation $\alpha_i=4$, so that $\sum_i\alpha_i^{-1}=1$. (iii) \emph{Gaussian baseline}: the exact linear-Gaussian posterior under the empirical mean and a rank-200 covariance of the training images, plus an isotropic term whose variance is that of the training images' residual outside the rank-200 subspace; it shows what access and a generic prior alone imply. Each system reports a 95\% interval for $q$ as estimate $\pm1.96\,\kappa$ times its spread (for EKI, the standard deviation of the ensemble's values of $q$, an ensemble-derived rather than an exact Bayesian interval). The factor $\kappa$ is set separately for each system and each decision quantity, on the calibration slices, so that coverage there is 95\%. This is an empirical adjustment of marginal coverage; it does not guarantee coverage conditional on an object, a perturbation or a shift in distribution. Before calibration every system was overconfident for the unreached quantity even on ordinary slices: raw coverage of $q$ was 0.685, 0.718 and 0.708 ($\kappa=2.08$, 2.32, 2.24), against 0.99, 0.985 and 0.74 for the within-reach control.

\paragraph{Diagnostics.} Nine ground-truth-free diagnostics: the standardised whitened total residual $T_0$, six angle-block residuals, latent typicality $\|\hat z\|^2$ (VAE systems only) and the lag-1 residual autocorrelation along the detector. They are combined in a Mahalanobis detector with mean, covariance and 5\% false-alarm threshold set on the calibration slices (empirical null): nine retained diagnostics (null covariance of rank 8) and $t=20.9$ (Laplace), $20.3$ (EKI); eight (rank 7) and $t=15.6$ for the Gaussian system, which has no typicality term (main study; other studies are in the deposited calibration files). On the test slices without errors, false alarms were 3.0\%, 3.5\% and 3.5\%, and coverage of $q$ was 0.93, 0.945 and 0.985; the Gaussian system's over-coverage lowers its own blind rate on ordinary slices.

\paragraph{Errors.} The error set has 240 members: inclusions of radius 2 or 3~px with amplitude $+0.3$, $+0.8$, $+1.5$ or $-0.15$ (a void; inclusions are clipped at zero, while the score uses the unclipped $\delta$, so for voids on near-zero bases the score is, if anything, too high), at six positions in each of four classes---\emph{in $R$} (centre within 4~px of $R$'s centre), \emph{adjacent} (centre 10--13~px from it), \emph{in $C$}, and \emph{elsewhere} within reach---and 48 transplants of radius-5 discs of real out-of-distribution content, 12 per class. Radius-5 transplants in the first two classes can extend up to 2~px across $R$'s boundary. In units of the prior standard deviation of $q$ ($\sigma_q=0.041$), inclusions in $R$ shift $q$ by a median of 0.5 (voids), 1.0, 2.8 and 5.2 (amplitudes $+0.3$, $+0.8$, $+1.5$), and transplants by 2.3; in units of the mean uninflated Laplace spread these are 1.2--11.6 and 5.1. All errors outside $R$ leave $q$ unchanged ($c^\top\delta=0$) except transplants that reach across its boundary. The largest perturbations are thus far outside the range of $q$ on ordinary slices (mean about 0.11).

\paragraph{Studies.} The \emph{main study} (Main) applies each error to five fixed test slices with two noise draws (10 trials per error). A robustness study, fixed before its trials, used 20 test slices per study with common random numbers (each slice run with and without each error under the same noise draw) and the \emph{excess} blind rate over the slice's own blindness as outcome: A1 repeats the main set-up; A2 moves the hidden disc and recalibrates; A3 uses attenuated access at $F=128{,}000$, with a central decision disc and the four classes replaced by depth rings by distance from the tube centre. Post-hoc analyses added A3 at $F=32{,}000$ and $512{,}000$ (10 slices each). A2 and A3 used 120 errors at alternate positions.

\paragraph{What was fixed when.} Table~\ref{tab:levels} summarises the hierarchy of evidence. Predictions and decision criteria of the main study were fixed, and the prediction file hashed (SHA-256), before any trial; the robustness study was fixed before its trials; seven post-hoc analysis plans were each written and approved before their own computation, after earlier results were known. The plans, hashes, deviations and failed configurations are in Appendix~\ref{app:protocol} and the deposited records. Uncertainty is reported with an error-only bootstrap (as pre-specified), a two-way bootstrap over errors and base slices, and a block bootstrap over errors and 10-slice CT blocks (5,000 replicates each); the last two are the relevant ones for generalisation beyond the base slices, and we quote two-way intervals unless stated otherwise (Table~\ref{tab:appB} compares all three for the differences in $\rho$). No interval resamples physical objects.

\begin{table}[t]\centering\small
\caption{Hierarchy of evidence. Each level was fixed at the time shown; later levels were designed with earlier results known.}\label{tab:levels}
\begin{tabular}{p{2.9cm}p{7.2cm}p{3.6cm}}\toprule
Level & Analyses & Fixed\\\midrule
Confirmatory & H1, H1b, H2 in the main study & before any trial\\
Pre-specified robustness & H1--H6 in A1--A3 & before each study's trials, after the main results\\
Post-hoc, specified before computation & seven analysis plans: (1) simpler predictors, bootstraps, attenuation strengths; (2) physics comparison at all strengths; (3) approximations, calibration, adjusted rankings, sensitivity; (4) components of the score; (5) trial-level analysis, linear-Gaussian reference, location control, prior--population check, $\kappa=1$, flagging; (6) detection-law dimension, leave-one-slice-out, flag budgets, physics at $\kappa=1$; (7) refitted and held-out-slice cross-validation & before each plan's computation, after earlier results\\
Exploratory & ties (Kendall $\tau_b$), within-region comparisons, descriptive figures & not fixed in advance\\\bottomrule
\end{tabular}
\end{table}

\section{Bounded reach}\label{sec:bounded}

\begin{figure}[t]\centering
\includegraphics[width=\linewidth]{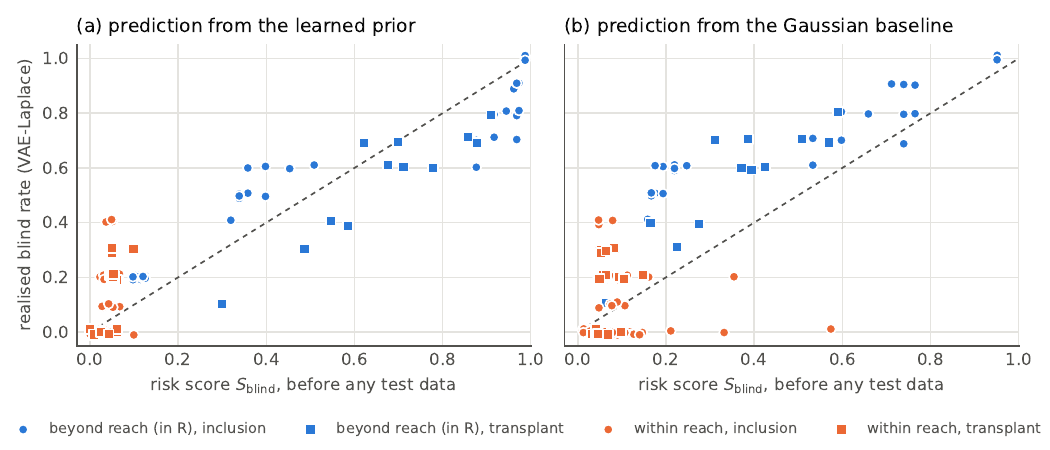}
\caption{Main study: blindness risk score $S_{\mathrm{blind}}$ (computed before any test data) against the realised blind rate of VAE-Laplace, one point per pre-specified error (small vertical jitter; realised rates are multiples of 0.1). (a) Learned score; (b) Gaussian prediction, which assigns lower scores than the learned system's observed blind rates beyond reach. Neither score is a calibrated probability.}\label{fig:scatter}
\end{figure}

\paragraph{Prospective tests.} The main study's criteria were: \textbf{H1}, Spearman $\rho$ between $S_{\mathrm{blind}}$ and the realised blind rate over the 240 errors $\geq0.5$, with an interval excluding zero; \textbf{H1b}, the same within $R$ (60 errors), $\rho\geq0.3$; \textbf{H2}, the learned score ranks realised blindness better than the Gaussian prediction. All were met (Table~\ref{tab:v1}, Fig.~\ref{fig:scatter}). Inclusions inside $R$ cannot move any diagnostic, since $A\delta=0$; they never raised the learned systems' detectors, and VAE-Laplace missed them confidently in 53\% of trials (transplants, 55\%). Within $R$, blindness rose with amplitude: realised 0.35, 0.71 and 0.87 for $+0.3$, $+0.8$, $+1.5$ (mean scores 0.22, 0.71, 0.95), and 0.20 for voids (mean score 0.09). In the reference H1b is guaranteed by design (Proposition~\ref{prop:lg}(ii)); in the realised systems such errors leave the data unchanged, so only their effect on $q$ varies, and within $R$ the Gaussian prediction ranks errors about as well as the learned one ($\rho=0.959$ against 0.948 for Laplace). The structural predictors alone would also have met H1's threshold (Table~\ref{tab:sens}), and H2 used the Gaussian prediction, which is a weak comparator under bounded reach. The prospective tests therefore establish that blindness can be ranked in advance, not that the learned score is needed to do it. The robustness study reproduced H1 and H2 in A1 and A2 for both systems (Appendix~\ref{app:protocol}).

\begin{table}[t]\centering\small
\caption{Main study: pre-specified decision criteria (95\% bootstrap intervals resampling errors).}\label{tab:v1}
\begin{tabular}{lccc}\toprule
Hypothesis & VAE-Laplace (primary) & VAE-EKI (secondary) & rule\\\midrule
H1: $\rho$(score, realised), 240 errors & 0.785 [0.701, 0.851] \pass & 0.844 [0.786, 0.887] \pass & $\geq0.5$, CI $>0$\\
H1b: $\rho$ within $R$, 60 errors & 0.948 [0.905, 0.968] \pass & 0.947 [0.910, 0.962] \pass & $\geq0.3$, CI $>0$\\
H2: $\rho$(VAE) $-$ $\rho$(Gaussian) & $+$0.334 [0.221, 0.453] \pass & $+$0.343 [0.229, 0.473] \pass & CI $>0$\\\bottomrule
\end{tabular}
\end{table}

\begin{table}[t]\centering\footnotesize
\caption{Comparison with simpler predictors (post-hoc; first and second analysis plans, and the fifth for $\Delta\rho$ under bounded reach). $\rho$: Spearman correlation with realised blindness (raw rate in Main, excess in A1--A3). Predictors: v1, the pre-specified score; v2, full-diagnostic variant; physics, physics-based baseline; Gauss, Gaussian prediction; $|c^\top\delta|$, effect size; region, indicator of the unreached region (A3: deepest ring). Brackets: two-way bootstrap (errors $\times$ base slices) 95\% intervals; Table~\ref{tab:appB} gives all three intervals for the differences in $\rho$, and the deposited analysis files give them for every quantity. ``Within reach'' excludes errors in the unreached region. n.d.: not defined (within reach at $F=512{,}000$ the EKI outcome is constant).}\label{tab:sens}
\setlength{\tabcolsep}{2.5pt}\resizebox{\linewidth}{!}{\begin{tabular}{llccccccc|ccc}\toprule
 & & \multicolumn{7}{c|}{all errors} & \multicolumn{3}{c}{within reach}\\
Study & System & v1 & v2 & physics & Gauss & $|c^\top\delta|$ & region & v1 $-$ physics [2-way] & v1 & physics & v1 $-$ physics [2-way]\\\midrule
Main & Laplace & 0.79 & 0.81 & 0.80 & 0.45 & 0.65 & 0.63 & $-$0.02 [$-$0.068, 0.061] & 0.62 & 0.66 & $-$0.04 [$-$0.141, 0.195]\\
 & EKI & 0.84 & 0.88 & 0.87 & 0.50 & 0.70 & 0.66 & $-$0.03 [$-$0.059, 0.002] & 0.72 & 0.78 & $-$0.06 [$-$0.126, 0.000]\\
A1 & Laplace & 0.60 & 0.56 & 0.59 & 0.44 & 0.59 & 0.56 & $+$0.01 [$-$0.114, 0.153] & 0.29 & 0.27 & $+$0.02 [$-$0.278, 0.410]\\
 & EKI & 0.70 & 0.74 & 0.74 & 0.37 & 0.57 & 0.52 & $-$0.04 [$-$0.076, 0.020] & 0.61 & 0.69 & $-$0.08 [$-$0.180, 0.149]\\
A2 & Laplace & 0.75 & 0.78 & 0.77 & 0.47 & 0.59 & 0.55 & $-$0.02 [$-$0.191, 0.196] & 0.64 & 0.69 & $-$0.04 [$-$0.391, 0.451]\\
 & EKI & 0.83 & 0.81 & 0.81 & 0.49 & 0.67 & 0.62 & $+$0.03 [$-$0.125, 0.199] & 0.75 & 0.69 & $+$0.05 [$-$0.280, 0.530]\\
A3, $F{=}128$k & Laplace & 0.47 & 0.58 & 0.68 & 0.52 & 0.55 & 0.57 & $-$0.21 [$-$0.491, 0.197] & 0.01 & 0.48 & $-$0.47 [$-$0.943, 0.369]\\
 & EKI & 0.75 & 0.71 & 0.44 & 0.63 & 0.59 & 0.65 & $+$0.32 [$-$0.101, 0.512] & 0.50 & $-$0.15 & $+$0.65 [$-$0.248, 1.024]\\
A3, $F{=}32$k & Laplace & 0.59 & 0.52 & 0.33 & 0.64 & 0.50 & 0.41 & $+$0.26 [$-$0.004, 0.418] & 0.47 & 0.01 & $+$0.46 [$-$0.039, 0.755]\\
 & EKI & 0.69 & 0.61 & 0.39 & 0.73 & 0.52 & 0.51 & $+$0.30 [0.033, 0.480] & 0.53 & 0.00 & $+$0.52 [0.012, 0.865]\\
A3, $F{=}512$k & Laplace & 0.58 & 0.58 & 0.54 & 0.59 & 0.66 & 0.63 & $+$0.04 [$-$0.001, 0.131] & 0.17 & $-$0.10 & $+$0.27 [0.000, 0.472]\\
 & EKI & 0.66 & 0.66 & 0.66 & 0.66 & 0.79 & 0.79 & 0.00 [$-$0.002, 0.001] & n.d. & n.d. & n.d.\\\bottomrule
\end{tabular}}
\end{table}

\paragraph{Comparison with simpler predictors.} In Main, A1 and A2 the learned score ranked realised blindness no better than the physics-based baseline, overall and within reach (Table~\ref{tab:sens}): v1 $-$ physics ranged from $-0.04$ to $+0.03$ overall, with every two-way interval including zero, and the full-diagnostic variant changed little. The region indicator alone reached $\rho=0.52$--0.66, and $|c^\top\delta|$ alone 0.57--0.70. Within reach the learned score's advantage over the Gaussian prediction ($\rho=0.62$ and 0.72 against $-0.03$ and 0.03 in Main) held against the Gaussian prediction only. That main-study value rests on five base slices whose own blind rate was 20\%; with 20 slices (A1) the base rate was 0.00 for Laplace and its within-reach criterion H4 just failed (0.29 against 0.3), so the within-reach result for Laplace is fragile.

\begin{table}[t]\centering\footnotesize
\caption{Trial-level analysis (fifth and seventh post-hoc plans; raw blind indicator of every trial). AUROC of the learned score (v1), the physics-based score and the region indicator. Calibration slope: logistic regression of the blind indicator on logit(score); calibration intercept: the same model with logit(score) as an offset (neither model has base-slice terms, and wide intervals fail to reject calibration rather than show it). Log-score gains: mean per-trial gain in 10-fold cross-validated log score from adding logit(v1) or logit(physics) to a structural model (region indicator and rank of $|c^\top\delta|$, with base-slice intercepts), and from adding v1 to the model that already contains physics. Folds hold out errors (all trials of an error, on every base slice), not base slices or objects; in each fold the intercepts and coefficients are refitted on the remaining errors, while the scores and the rank of $|c^\top\delta|$ are computed from the model alone. The gains therefore measure incremental information for unseen errors on the same base slices. Calibration models are fitted to all trials, not cross-validated. Brackets: two-way bootstrap 95\% intervals; for log-score gains every model is refitted in every fold of every replicate (seventh plan, 5,000 replicates). Table~\ref{tab:H} gives sensitivity to the fold partition and cross-validation on held-out base slices. $^\ddagger$n.i.: not interpretable. At $F=512{,}000$, where blindness is confined almost entirely to the deepest ring, the refitted intervals are orders of magnitude wider than elsewhere (for example $[-5.45, 0.01]$ for EKI, v1 over structure), consistent with refitted models that separate, or nearly separate, the outcomes in some replicates and give a few held-out trials extreme log scores; the values are in the deposited results.}\label{tab:trial}
\setlength{\tabcolsep}{2.5pt}\resizebox{\linewidth}{!}{\begin{tabular}{llcccccccc}\toprule
 & & \multicolumn{3}{c}{AUROC} & \multicolumn{2}{c}{calibration (v1)} & \multicolumn{3}{c}{log-score gain per trial}\\\cmidrule(lr){3-5}\cmidrule(lr){6-7}\cmidrule(lr){8-10}
Study & System & v1 & physics & region & slope & intercept & v1 over structure & physics over structure & v1 over physics\\\midrule
Main & Laplace & 0.78 & 0.79 & 0.70 & 0.58 [0.37, 1.85] & $+$1.03 [$-$2.07, 2.84] & $+$0.077 [0.001, 0.181] & $+$0.077 [0.001, 0.186] & $-$0.002 [$-$0.025, 0.016]\\
 & EKI & 0.81 & 0.81 & 0.73 & 0.74 [0.51, 2.63] & $+$1.15 [$-$1.61, 2.97] & $+$0.061 [$-$0.009, 0.169] & $+$0.062 [$-$0.005, 0.175] & $-$0.001 [$-$0.071, 0.010]\\
A1 & Laplace & 0.95 & 0.94 & 0.87 & 1.14 [0.93, 1.51] & $-$0.87 [$-$1.45, $-$0.38] & $+$0.077 [0.036, 0.146] & $+$0.078 [0.039, 0.146] & $-$0.002 [$-$0.022, 0.005]\\
 & EKI & 0.93 & 0.93 & 0.85 & 1.12 [0.81, 1.99] & $-$0.28 [$-$1.19, 0.63] & $+$0.073 [$-$0.015, 0.168] & $+$0.073 [$-$0.021, 0.169] & 0.000 [$-$0.019, 0.002]\\
A2 & Laplace & 0.83 & 0.84 & 0.75 & 0.72 [0.56, 1.06] & $+$0.21 [$-$1.04, 1.16] & $+$0.089 [0.034, 0.405] & $+$0.089 [0.033, 0.404] & $-$0.001 [$-$0.019, 0.013]\\
 & EKI & 0.86 & 0.85 & 0.78 & 0.81 [0.59, 1.32] & $+$0.44 [$-$0.70, 1.33] & $+$0.071 [0.024, 0.345] & $+$0.070 [0.025, 0.348] & 0.000 [$-$0.024, 0.013]\\
A3, $F{=}128$k & Laplace & 0.66 & 0.68 & 0.63 & 0.36 [0.20, 0.61] & $+$0.21 [$-$1.27, 1.32] & $+$0.018 [0.002, 0.041] & $+$0.013 [$-$0.002, 0.033] & $+$0.004 [$-$0.007, 0.026]\\
 & EKI & 0.77 & 0.73 & 0.71 & 0.58 [0.38, 0.95] & $+$0.41 [$-$0.84, 1.41] & $+$0.045 [0.013, 0.084] & $+$0.028 [0.001, 0.060] & $+$0.024 [$-$0.002, 0.067]\\
A3, $F{=}32$k & Laplace & 0.68 & 0.62 & 0.61 & 0.31 [0.14, 0.76] & $-$0.35 [$-$2.36, 1.46] & $+$0.009 [$-$0.029, 0.046] & $-$0.001 [$-$0.022, 0.029] & $+$0.020 [$-$0.016, 0.047]\\
 & EKI & 0.70 & 0.63 & 0.62 & 0.34 [0.14, 0.76] & $+$0.02 [$-$1.81, 1.62] & $+$0.017 [$-$0.018, 0.060] & $-$0.001 [$-$0.020, 0.027] & $+$0.035 [$-$0.002, 0.096]\\
A3, $F{=}512$k & Laplace & 0.98 & 0.97 & 0.90 & 2.72 [1.20, 18.27] & $-$2.42 [$-$3.20, $-$1.79] & n.i.$^\ddagger$ & n.i.$^\ddagger$ & n.i.$^\ddagger$\\
 & EKI & 0.99 & 0.99 & 0.92 & 2.44 [1.60, 7.77] & $-$1.67 [$-$2.39, $-$1.08] & n.i.$^\ddagger$ & n.i.$^\ddagger$ & n.i.$^\ddagger$\\\bottomrule
\end{tabular}}
\end{table}

\paragraph{Trial-level analysis.} Spearman $\rho$ over a designed error set depends on how the set was built, and each realised rate rests on 10 or 20 trials. We therefore also modelled the blind indicator of every trial (Table~\ref{tab:trial}). Under bounded reach the learned and physics-based scores had the same discrimination (AUROC 0.78--0.95, within 0.01 of each other), above the region indicator (0.70--0.87). Each improved the cross-validated log score of a structural model (region indicator and rank of $|c^\top\delta|$, with base-slice intercepts) by 0.06--0.09 nats per trial. The intervals in Table~\ref{tab:trial} refit every model in every fold of every bootstrap replicate (seventh plan). With them, these gains were above zero in four of the six bounded-reach pairs but not for EKI in Main and A1, where intervals that did not refit had excluded zero; the change is marginal (lower limits moved from $+0.005$--$+0.023$ to $-0.005$--$-0.021$). Neither score improved detectably on the other: the gain from adding the learned score to a model that already contains the physics-based one was between $-0.002$ and 0.000 in every bounded-reach pair, with refitted upper limits of at most 0.016 nats per trial, and, by the rule fixed in advance, no incremental improvement from the learned score beyond the physics-based score was detected in any of the twelve study--system pairs without refitting, or in any of the ten whose refitted intervals are interpretable (nor from physics beyond the learned score). At $F=512{,}000$ the refitted intervals are orders of magnitude wider than elsewhere and are marked not interpretable: blindness there is confined almost entirely to the deepest ring, and the refitted models appear to separate, or nearly separate, the outcomes in some replicates, giving a few held-out trials extreme log scores (Table~\ref{tab:trial}). Across 101 random partitions of the errors into folds, this gain varied by at most 0.006 under bounded reach, well within its interval. No equivalence margin was fixed in advance, so this is an absence of detected improvement, not a demonstration that the two scores carry the same information; descriptively, the refitted upper limits are at most about a fifth of either score's gain over the structural model. The folds hold out errors, not base slices or objects: all trials of an error, on every base slice, are left out together, and the base-slice intercepts and coefficients are refitted on the remaining errors in each fold. Unlike the prospective score, these models are fitted to realised outcomes. The gains therefore measure the incremental information each score carries for unseen errors on the same base slices; they do not estimate performance on new objects without outcome-based fitting. The seventh plan also held out base slices, grouped by CT block, together with the errors, replacing the slice intercepts by a common one (Table~\ref{tab:H}). The learned score again added no detectable information beyond physics in any pair, and either score improved the structural model with an interval above zero for A1 Laplace and A2 only. In the main study five slices were too few for this design: the structural model predicted unseen slices poorly (mean log score $-0.94$ and $-2.66$ nats per trial, against $-0.23$ and $-0.13$ with slice intercepts), and the intervals were very wide. Held-out slices come from the same sample tubes as the training slices, so this is not validation on independent objects. The calibration slope and intercept of the learned score (models without base-slice terms, fitted to all trials) had intervals containing 1 and 0 in five of the six bounded-reach pairs, the exception being A1 Laplace. With five base slices these intervals are wide (slope [0.37, 1.85] for Main Laplace); they fail to reject calibration rather than show it, and calibration-in-the-large remains $-0.08$ to $-0.09$ in the main study (Section~\ref{sec:explain}).

\begin{table}[t]\centering\footnotesize
\caption{Rankings after adjustment for properties of the errors (third post-hoc plan). Partial $\rho$: all errors, adjusted for the rank of $|c^\top\delta|$ and the unreached-region indicator. Stratified $\rho$: within strata of location class $\times$ error kind $\times$ amplitude. $\Delta$: v1 $-$ physics. Brackets: two-way bootstrap 95\% intervals; $^*$: interval excludes zero. The within-reach analysis with a distance spline is in Table~\ref{tab:adjdist}.}\label{tab:adj}
\setlength{\tabcolsep}{2.5pt}\resizebox{\linewidth}{!}{\begin{tabular}{llcccccc}\toprule
 & & \multicolumn{3}{c}{partial $\rho$ (all errors)} & \multicolumn{3}{c}{stratified $\rho$}\\\cmidrule(lr){3-5}\cmidrule(lr){6-8}
Study & System & v1 & physics & $\Delta$ & v1 & physics & $\Delta$\\\midrule
Main & Laplace & $+$0.58$^*$ [0.009, 0.686] & $+$0.62 [$-$0.037, 0.730] & $-$0.04 [$-$0.141, 0.135] & $+$0.34$^*$ [0.018, 0.517] & $+$0.34$^*$ [0.071, 0.497] & 0.00 [$-$0.198, 0.213]\\
 & EKI & $+$0.66$^*$ [0.005, 0.740] & $+$0.73 [0.000, 0.780] & $-$0.06 [$-$0.127, 0.015] & $+$0.36$^*$ [0.095, 0.527] & $+$0.39$^*$ [0.138, 0.537] & $-$0.03 [$-$0.122, 0.020]\\
A1 & Laplace & $+$0.26$^*$ [0.053, 0.332] & $+$0.25 [$-$0.080, 0.382] & $+$0.01 [$-$0.229, 0.323] & $+$0.30$^*$ [0.050, 0.509] & $+$0.23 [$-$0.005, 0.459] & $+$0.07 [$-$0.149, 0.281]\\
 & EKI & $+$0.51 [$-$0.017, 0.594] & $+$0.58 [$-$0.032, 0.659] & $-$0.07 [$-$0.153, 0.060] & $+$0.36$^*$ [0.127, 0.555] & $+$0.34$^*$ [0.116, 0.543] & $+$0.02 [$-$0.115, 0.158]\\
A2 & Laplace & $+$0.58$^*$ [0.178, 0.690] & $+$0.65 [$-$0.001, 0.819] & $-$0.08 [$-$0.355, 0.284] & $+$0.18 [$-$0.213, 0.541] & $+$0.35 [$-$0.080, 0.632] & $-$0.18 [$-$0.589, 0.332]\\
 & EKI & $+$0.67$^*$ [0.123, 0.769] & $+$0.66 [$-$0.088, 0.846] & $+$0.01 [$-$0.271, 0.353] & $+$0.31 [$-$0.098, 0.565] & $+$0.24 [$-$0.208, 0.572] & $+$0.07 [$-$0.296, 0.425]\\
A3, $F{=}128$k & Laplace & $+$0.04 [$-$0.325, 0.430] & $+$0.47 [$-$0.088, 0.653] & $-$0.42 [$-$0.905, 0.422] & $+$0.08 [$-$0.285, 0.455] & $+$0.12 [$-$0.259, 0.447] & $-$0.03 [$-$0.502, 0.464]\\
 & EKI & $+$0.56$^*$ [0.054, 0.657] & $-$0.13 [$-$0.494, 0.410] & $+$0.68 [$-$0.215, 0.993] & $+$0.39 [$-$0.009, 0.610] & $+$0.14 [$-$0.289, 0.495] & $+$0.24 [$-$0.268, 0.682]\\
A3, $F{=}32$k & Laplace & $+$0.38$^*$ [0.063, 0.557] & $-$0.08 [$-$0.382, 0.352] & $+$0.46 [$-$0.010, 0.697] & $+$0.19 [$-$0.178, 0.452] & $+$0.05 [$-$0.332, 0.359] & $+$0.14 [$-$0.226, 0.576]\\
 & EKI & $+$0.52$^*$ [0.212, 0.628] & $-$0.04 [$-$0.346, 0.311] & $+$0.56$^*$ [0.071, 0.821] & $+$0.19 [$-$0.189, 0.481] & $-$0.05 [$-$0.390, 0.320] & $+$0.24 [$-$0.222, 0.680]\\
A3, $F{=}512$k & Laplace & $+$0.12$^*$ [0.020, 0.241] & $-$0.02 [$-$0.082, 0.033] & $+$0.14$^*$ [0.016, 0.311] & $+$0.17 [$-$0.159, 0.459] & $+$0.08 [$-$0.263, 0.444] & $+$0.09 [$-$0.025, 0.372]\\
 & EKI & $+$0.07$^*$ [0.016, 0.138] & $+$0.01 [0.000, 0.042] & $+$0.05$^*$ [0.010, 0.115] & $+$0.35$^*$ [0.101, 0.524] & $+$0.36$^*$ [0.106, 0.541] & $-$0.01$^*$ [$-$0.039, $-$0.002]\\\bottomrule
\end{tabular}}
\end{table}

\paragraph{Rankings after adjustment.} Two measures adjust for properties of the errors and answer different questions (Table~\ref{tab:adj}). The \emph{stratified} $\rho$ asks whether a predictor ranks errors within groups of similar errors (strata of location class $\times$ error kind $\times$ amplitude; ranks centred within strata, Pearson correlation of the pooled centred ranks; strata with fewer than three errors dropped). The \emph{partial} $\rho$ converts prediction and outcome to ranks, regresses each by least squares on an intercept, the rank of $|c^\top\delta|$ and the region indicator, and correlates the residuals; it adjusts for one binary feature of location only. Both use average ranks for ties and are recomputed, regressions included, in every bootstrap replicate. Under bounded reach the learned and physics-based scores showed no detectable difference on either measure: v1 $-$ physics was $-0.08$ to $+0.01$ in partial $\rho$ and $-0.18$ to $+0.07$ in stratified $\rho$, with every two-way interval including zero. Counted separately against the rule fixed in the third plan (stratified or partial interval excluding zero), the learned score met this rule for information beyond these variables in all six bounded-reach pairs and the physics-based baseline in three, but this asymmetry rests on lower interval limits that differ by hundredths while the point estimates are nearly equal or favour physics; it is not evidence of a difference. The Gaussian prediction did not (partial $\rho\leq0.13$). The main-study intervals are strongly skewed: leaving out one of its five base slices lowered the partial $\rho$ of both scores to near zero (0.08 and 0.01 for Laplace, 0.01 and 0.02 for EKI), while leaving out any other slice changed little and the stratified $\rho$ stayed at 0.25--0.42; one slice drives the main-study partial $\rho$. To control location within the reached region, the fifth plan added a natural cubic spline in the distance to the unreached disc (Table~\ref{tab:adjdist}): within reach, the two scores' point estimates were again nearly identical (for example 0.69 and 0.69 for Main Laplace) and their difference included zero everywhere, while the learned score's interval excluded zero in five of six pairs and the physics-based baseline's in one. At $F=512{,}000$ almost all within-reach outcomes are zero, and these values are not interpretable (Table~\ref{tab:adjdist}). The two scores agree closely but not perfectly: their rank correlation across errors is 0.84 under bounded reach (0.62--0.64 within reach) and 0.42--0.59 under attenuation, so their similar trial-level performance is an empirical finding rather than a consequence of identical scores. These are statistical findings, not a mechanism. The physics-based baseline's only remaining input after adjustment is $\|A_w\delta\|^2$, the error's detectability in the whitened measurements, which suggests that detectability carries the residual information, but its wider intervals do not let us say that it carries all of it.

\paragraph{Access explains much of the blind set.} The Gaussian \emph{system} was itself confidently wrong for errors beyond reach in 44\% (Main), 30\% (A1), 28\% (A2) and 40\% (A3) of trials. Under bounded reach, which errors are blind is predicted by access, the measurement physics and the decision quantity; the learned prior mainly sets how often blindness occurs on ordinary images, through the overconfidence that calibration had to correct.

\section{Attenuated access}\label{sec:atten}

\begin{figure}[t]\centering
\includegraphics[width=\linewidth]{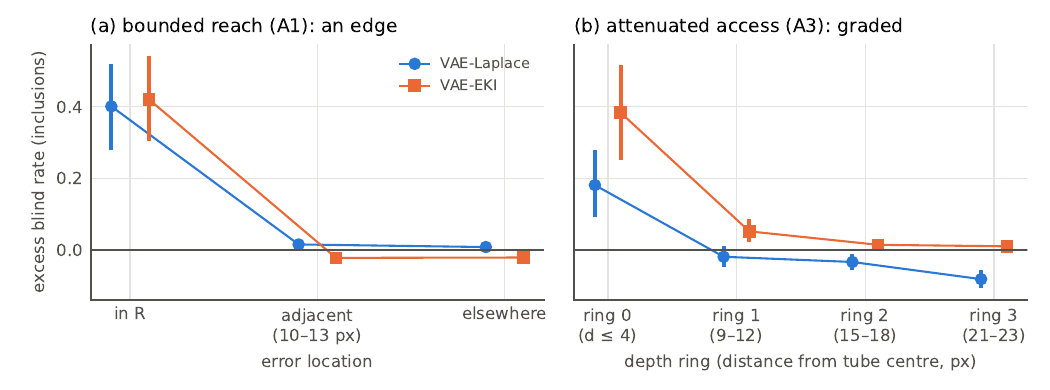}
\caption{Shape of the blind set (inclusions; excess blind rate, mean and 95\% bootstrap interval resampling errors). (a) Bounded reach (A1): an edge at the boundary of the unreached region. (b) Attenuated access (A3, $F=128{,}000$): blindness concentrated in the deepest ring and graded outward. Transplants are omitted because their contrast against the base varies from slice to slice. The shapes depend on the chosen decision quantity, the error geometry and locations, and the measurement configuration.}\label{fig:shape}
\end{figure}

\paragraph{Shape and calibration.} Bounded reach produced an edge: in A1, excess blindness was $+0.40$ inside $R$ and close to zero just outside it ($+0.02$ Laplace, $-0.02$ EKI; Fig.~\ref{fig:shape}a). Attenuated access at $F=128{,}000$ produced a graded profile, highest in the deepest ring ($+0.18$ Laplace, $+0.38$ EKI) with residual detection (0.04 for Laplace inclusions), and falling outward (Fig.~\ref{fig:shape}b). Because an error that raises an alarm can also hide the base's own blindness, excess can be negative; the rate of newly blind trials fell monotonically for Laplace (0.24, 0.07, 0.03, 0.00 across rings 0--3) and from ring 0 to rings 1--3 for EKI (0.39, 0.08, 0.03, 0.05). The shape depended on the strength $F$ (Table~\ref{tab:b4}): graded at $F=32{,}000$ (EKI $+0.23$, $+0.11$, $+0.03$, $-0.01$ from ring~0 outward), confined to the deepest ring at $512{,}000$ ($+0.29$ Laplace, $+0.38$ EKI; one blind trial in 900 outside it for Laplace, none for EKI). Calibration moved the same way: raw coverage of central $q$ was 0.98 at $F=32{,}000$, about 0.95 at $128{,}000$ and 0.92--0.95 at $512{,}000$ ($\kappa=0.76$--0.81, 0.92--0.98 and 1.03--1.18), against about 0.7 under bounded reach. Weak access kept the reported uncertainty honest on average; no access did not.

\begin{table}[t]\centering\footnotesize
\caption{Attenuation-strength sensitivity (post-hoc; thresholds shown but not used as pass/fail). Excess blind rate; 95\% bootstrap intervals resampling errors. $F=128{,}000$ is the pre-specified A3 study (20 base slices; as in Table~\ref{tab:add}); the neighbouring strengths use 10 base slices each.}\label{tab:b4}
\setlength{\tabcolsep}{3.5pt}\begin{tabular}{llcccc}\toprule
$F$ & System & H1: $\rho\geq0.5$ & H2: $\Delta\rho$ (VAE $-$ Gauss) & H6: ring 0 $-$ ring 1 & H6: ring 1 $-$ ring 3\\\midrule
$32{,}000$ & Laplace & 0.587 [0.427, 0.713] & $-$0.054 [$-$0.154, 0.044] & $+$0.083 [0.000, 0.171] & $+$0.046 [$-$0.004, 0.100]\\
 & EKI & 0.688 [0.564, 0.786] & $-$0.041 [$-$0.130, 0.045] & $+$0.117 [0.021, 0.221] & $+$0.125 [0.075, 0.175]\\
\midrule
$128{,}000$ & Laplace & 0.466 [0.277, 0.624] & $-$0.052 [$-$0.161, 0.047] & $+$0.200 [0.104, 0.304] & $+$0.062 [0.025, 0.100]\\
 & EKI & 0.754 [0.632, 0.847] & $+$0.120 [0.015, 0.238] & $+$0.331 [0.200, 0.469] & $+$0.042 [0.006, 0.079]\\
\midrule
$512{,}000$ & Laplace & 0.578 [0.450, 0.683] & $-$0.009 [$-$0.028, 0.000] & $+$0.283 [0.133, 0.438] & $+$0.004 [0.000, 0.013]\\
 & EKI & 0.662 [0.539, 0.750] & 0.000 [0.000, 0.000] & $+$0.379 [0.221, 0.542] & 0.000 [0.000, 0.000]\\
\bottomrule\end{tabular}\end{table}

\paragraph{What predicts blindness.} At $F=128{,}000$ the pre-specified score met H1 and H2 for EKI but not for Laplace (Appendix~\ref{app:protocol}). Across strengths (Table~\ref{tab:sens}), at the weakest attenuation, $F=32{,}000$, the learned score ranked realised blindness better than the physics-based baseline for EKI (v1 $-$ physics $+0.30$ [0.033, 0.480]); for Laplace the interval reached zero ($+0.26$ [$-0.004$, 0.418]), and the physics-based baseline carried almost no information within reach ($\rho\approx0.01$). After adjustment for effect size and region, the learned score again ranked better than physics for EKI at this strength (partial $\rho$ difference $+0.56$ [0.071, 0.821]), consistent with the unadjusted result; at $512{,}000$ three adjusted differences exclude zero, with mixed sign (partial $\rho$: $+0.14$ Laplace, $+0.05$ EKI; stratified $\rho$: $-0.01$ EKI; Table~\ref{tab:adj}). This was not specific to the learned prior: the Gaussian prediction ranked realised blindness at least as well (0.64 and 0.73 against 0.59 and 0.69), and at no strength did the learned score beat it with a two-way interval excluding zero (at $F=128{,}000$, where EKI met H2 with an error-only interval of [0.015, 0.238], the two-way interval was [$-0.064$, 0.234]). At $512{,}000$, where blindness was confined to the deepest ring, all predictions ranked errors alike (EKI: 0.66 each) and the region indicator did better (0.79). At the trial level (Table~\ref{tab:trial}) the learned score improved the structural model at $F=128{,}000$ for both systems but not at $32{,}000$ (at $512{,}000$ the refitted log-score intervals are not interpretable), and nowhere did it add detectable information beyond the physics-based score (at $F=32{,}000$ the gains were $+0.020$ and $+0.035$ nats per trial, with intervals including zero); its AUROC exceeded the physics-based score's by 0.05--0.07 at $F=32{,}000$ and by 0.04 for EKI at $128{,}000$, with intervals reaching down to about zero. Following the rule fixed in the second plan, we describe a learned-prior contribution under attenuation as open rather than absent. Rank correlation across errors and trial-level log score measure different things: the first, how well errors are ordered by blind rate; the second, how well individual trials are predicted. At the weakest attenuation prior information improved the ordering of errors, but this did not appear as a detectable gain in probabilistic prediction beyond the physics-based score.

\paragraph{Taking calibration out of the comparison.} Because each system's $\kappa$ was tuned separately, the fifth plan recomputed realised blindness and both predictions with $\kappa=1$ for every system. Under bounded reach the learned score then ranked realised blindness better than the Gaussian prediction by $+0.32$ to $+0.44$ (all intervals excluding zero); under attenuation the two did not differ (for EKI at $F=128{,}000$, $+0.09$ [$-0.06$, 0.19], against $+0.12$ with calibrated intervals). Against the physics-based baseline with $\kappa=1$ (sixth plan) the learned score ranked no better: v1 $-$ physics ranged from $-0.27$ to $+0.17$, physics was ahead in most pairs, and no two-way interval excluded zero except a negligible one at $F=512{,}000$ for EKI ($-0.001$).

\section{What explains the ranking}\label{sec:explain}

\paragraph{The reference, numerically.} Under the linearised model at the same 100 points, with objects drawn from the prior and the detector's null recalibrated on error-free draws (false alarms 5\%), the joint and product probabilities differed by at most 0.002 for every error, and Monte Carlo and analytic non-coverage at $\kappa=1$ by at most 0.006, in every study. Since Proposition~\ref{prop:lg} guarantees this agreement, the check verifies the implementation rather than providing independent evidence. An earlier Monte Carlo version (third plan), which drew only noise around the linearisation point (no variation of the object and no shrinkage bias) and inherited a detector null that the linearised diagnostics did not reproduce (23--45\% false alarms on error-free draws), ranked realised blindness worse than $S_{\mathrm{blind}}$. The prior-averaged version with the recalibrated null showed no detectable difference from it under bounded reach (differences $-0.16$ to $-0.01$, every two-way interval including zero). This is consistent with the deficit arising from the noise-only simulation and the mismatched null, which met the rule fixed in the fifth plan; but the two changes were made together, so it cannot be assigned to either alone, and no detected difference is not equivalence. Under attenuation the prior-averaged version still ranked worse at $F=32{,}000$ ($-0.27$ and $-0.25$; interval excluding zero for Laplace) and alike at $512{,}000$. A first computation of this check inverted a singular null covariance (the total-residual diagnostic is a linear combination of the six block residuals); the corrected computation uses a pseudo-inverse, as the realised detector does, and both are logged in the deposited records.

\paragraph{Components and dependence.} The fourth plan compared each component with its realised counterpart (Appendix~\ref{app:more}). The analytic detection component ranked realised alarm rates in every study and system ($\rho=0.54$--0.94) and was within $\pm0.05$ of them on average in 5 of 12 pairs; a Monte Carlo version ranked them as well but exceeded them by 0.32--0.54. The calibrated posterior spread described realised non-coverage better than the sampling variability of the linearised estimator in 9 of 12 pairs, as Proposition~\ref{prop:lg} leads one to expect for rates averaged over slices. In the realised systems, per-error blind rates differed little from the product of their realised non-detection and non-coverage rates (mean difference $-0.021$ to $+0.009$), but not little enough in ranking to meet the rule fixed in advance for calling the events independent ($\rho$ below 0.95 in 6 of 12 pairs). Swapping one component at a time, a sensitivity analysis rather than a causal decomposition, showed that neither component alone accounts for the analytic score's advantage over the Monte Carlo version: the coverage component mattered under bounded reach and the detection component under attenuation.

\begin{figure}[t]\centering
\includegraphics[width=\linewidth]{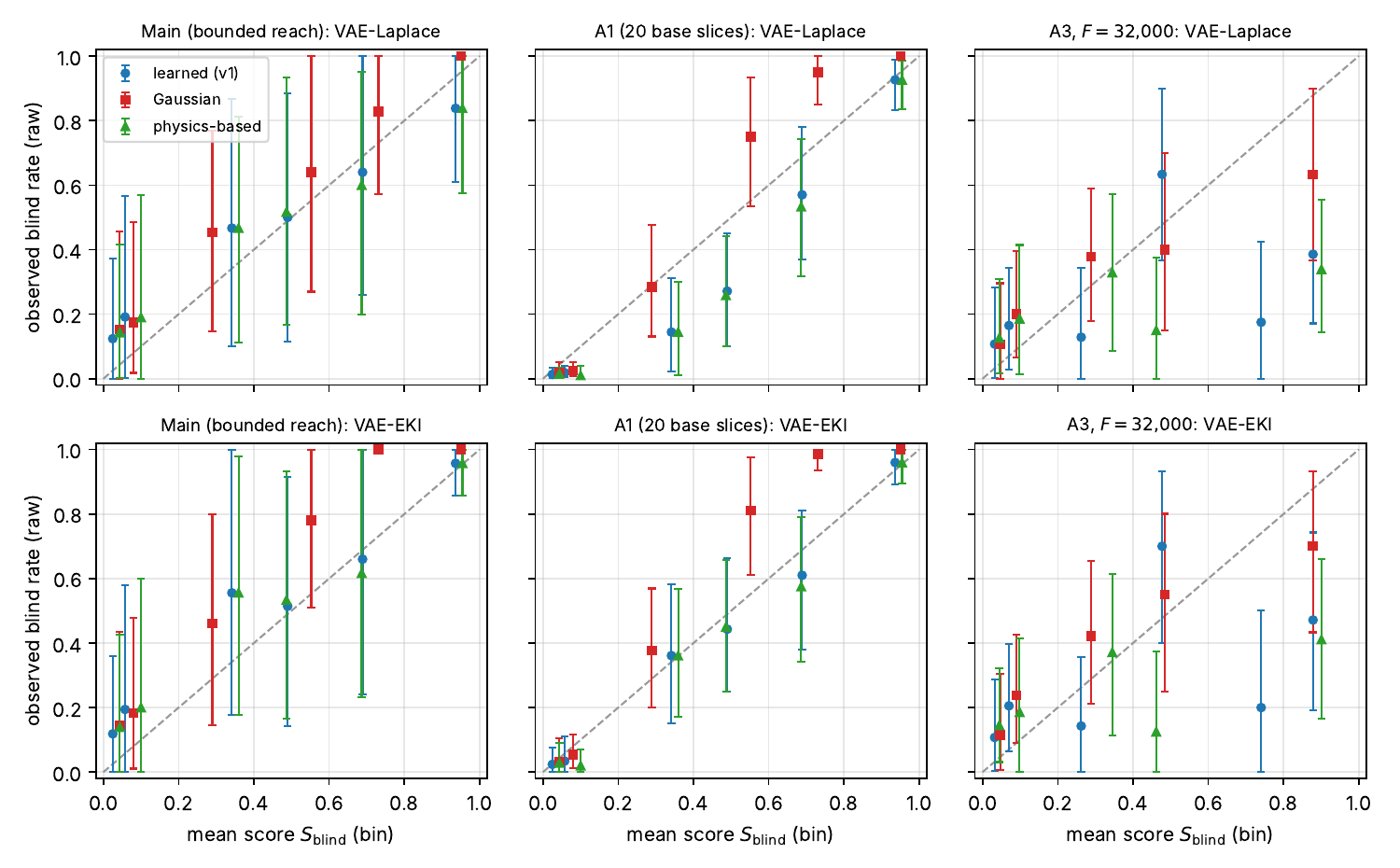}
\caption{Reliability of the scores against the raw blind rate (third post-hoc plan), for the learned (v1), Gaussian and physics-based predictions; bins of score, observed rate with two-way bootstrap 95\% intervals; dashed line: perfect calibration.}\label{fig:rel}
\end{figure}

\paragraph{Calibration.} $S_{\mathrm{blind}}$ is not a calibrated probability (Fig.~\ref{fig:rel}). Its calibration-in-the-large (mean score minus observed rate; Table~\ref{tab:D}), which checks only the average level, ranged from $-0.09$ (main study) to $+0.13$ (strongest attenuation), outside the $\pm0.05$ fixed in advance. At the trial level its calibration slope was 0.31--0.58 at $F=32{,}000$ and $128{,}000$ (scores too spread out) and 2.4--2.7 at $512{,}000$ (too compressed). Its Brier score showed no detectable difference from the physics-based baseline's under bounded reach (differences $-0.0007$ to $+0.0001$, every two-way interval including zero; deposited results of the third plan). Its Brier differences from the Gaussian prediction, negative under bounded reach and positive under attenuation, are in Table~\ref{tab:D}.  Part of the level error comes from the detection law: with $\nu=9$ at the empirical thresholds, the $\chi^2$ law implies false-alarm rates of 1.2--2.2\% under bounded reach and 6--9\% under attenuation rather than the empirical 5\%, shifting non-detection by a roughly constant amount. Recomputing the score with $\nu$ equal to the rank (8) and $t$ the 95\% quantile of $\chi^2_8$ (sixth plan) changed $\rho$ by $-0.03$ to $+0.03$, every interval including zero, and calibration-in-the-large by at most 0.01; the ranking is insensitive to this choice.

\paragraph{Population and linearisation points.} The test slices lie far outside the typical set of the latent prior: the squared norm of their encoder means averaged 66.5 (5--95\%: 30.8--104.4), against 32 (20.1--46.2) under $\chi^2_{32}$ (Kolmogorov--Smirnov distance 0.80), consistent with the KL weight reaching 1 only at the end of training. $S_{\mathrm{blind}}$ therefore averages local sensitivities over objects unlike the test population, which may contribute to its miscalibration. Strong shrinkage of such objects towards the prior is also a plausible cause of the systems' overconfidence for the unreached quantity, which required $\kappa\approx2$ (Section~\ref{sec:design}); we have not tested this. Its ranking was nonetheless stable: 25 or 50 linearisation points instead of 100 changed $\rho$ by at most 0.05, and points taken from encoded training images by $-0.09$ to $+0.02$. Stability shows that the ranking does not hinge on a few points or on where they are drawn; it does not show that the linearisations reproduce the non-linear systems. One reading, consistent with the physics-based baseline's performance, is that the ranking is dominated by quantities that change little across $z_k$: the whitened operator acting on $\delta$ and the decision functional. Scaling $\kappa$ from 0.75 to 1.5 times its calibrated value lowered the blind rate in $R$ from about 0.65 to 0.40 in the main study while $\rho$ stayed between 0.73 and 0.85. A physics baseline with no VAE-derived input ranked realised blindness lower than the physics-based baseline by 0.14--0.19 in five of the six bounded-reach study--system pairs and 0.02 higher in the sixth (A1, Laplace; Table~\ref{tab:D}), with every two-way interval for the difference including zero; by the rule fixed in the third plan we attribute the bounded-reach agreement to measurement structure, but without the model-error whitening every within-reach error is deemed certainly detected, so the within-reach part depends on modelling the decoder's error.

\paragraph{A use of the score.} Before any target data are seen, one can flag the errors the system cannot vouch for. In every study a quarter of the errors lie in the unreached region (or deepest ring), so a budget of 25\% coincides with the design: flagging the top quarter by $S_{\mathrm{blind}}$ captured 55--91\% of blind trials under bounded reach and 45--100\% under attenuation, with precision (the blind rate among flagged trials) 0.43--0.59 and 0.30--0.54, and the region indicator gave nearly the same flags. Across budgets from 5\% to 50\% (sixth plan), precision fell from 0.92--1.00 at 5\% to 0.22--0.39 at 50\% under bounded reach, and the learned flag captured more blind trials than the physics-based flag in none of 120 budget $\times$ study--system combinations. These capture and precision values are conditional on the constructed error set: the share of errors placed in the unreached region, and with it the prevalence of blind errors, was fixed by design. They show that the score can prioritise errors within this setting, not how well it would detect defects in operation, which would need a model of the errors expected in deployment. Its value here comes from access and effect size.

\section{Discussion and limitations}\label{sec:disc}

\emph{(a) Blind errors can be predicted prospectively:} supported in the tested setting. A score computed before the target data ranked which errors learned non-linear inference missed without warning, in two bounded-reach geometries for both learned systems and under attenuated access for the ensemble system. The pre-specified tests do not show that anything beyond access and effect size is needed for this. \emph{(b) Local linearisation of learned inference provides an effective predictor:} qualified support. The score approximates an exact linear-Gaussian reference, ranked blindness well and identified detectable errors, but it is not a calibrated probability, it averages over a prior that does not describe the test population, and an exact Monte Carlo evaluation of the local model showed no detectable difference from the score under bounded reach and ranked worse at the weakest attenuation. \emph{(c) Learned-prior geometry contributes information that simpler predictors lack:} not established. Under bounded reach the learned score showed no detectable advantage over the physics-based baseline in rankings, adjusted rankings or trial-level log scores; with no equivalence margin fixed in advance, this does not show that the two carry identical information. At the weakest attenuation studied, prior information improved rank correlations over the physics-based baseline, but a Gaussian prior did as well, and the improvement in ranking did not appear as a detectable gain in trial-level log score beyond physics. Whether a learned prior adds stable structure where data and prior trade off continuously remains open.

\paragraph{What is established, approximated and observed.} Three levels of statement should be kept apart. \emph{Exact:} for a correctly specified linear-Gaussian model averaged over the prior, non-detection and non-coverage are independent and the risk factorises (Proposition~\ref{prop:lg}). \emph{Approximate:} the score applies this product to local linearisations of a non-linear system, with an inflation factor fitted for coverage, a one-component detection law and an average over prior draws (Section~\ref{sec:predict}); it is checked numerically only against its own linearised model. \emph{Empirical:} in the realised systems, per-error blind rates were close on average to the product of realised non-detection and non-coverage, but not close enough in ranking to meet the rule fixed for calling the events independent, and trial-level odds ratios were imprecise (Section~\ref{sec:explain}); independence in the realised systems is neither established nor excluded.

The contribution is therefore an empirical and methodological investigation of which information suffices to anticipate silent failure, supported by an exact Gaussian reference; it is neither a general theory of uncertainty failure nor a demonstration that learned priors predict it better. Much of the structure of the blind set follows from measurement access and the decision quantity, as Proposition~\ref{prop:lg} predicts for the null space; prior information improved rankings at the weakest attenuation studied. Across the three attenuation strengths we observed a consistent pattern, which we present as an observation in this setting: as access weakened towards none, the blind set changed from a graded profile to an edge, the reported uncertainty beyond reach went from honest on average to overconfident, and the learned, Gaussian and physics-based predictions converged as blindness became dominated by location. The Laplace approximation was the weaker system throughout.

\paragraph{Calibration and the practical message.} Interval inflation was already set per decision quantity, so the unreached quantity had its own $\kappa$; even so, every system needed $\kappa\approx2$--2.3 there on ordinary slices. Region-wise or conformal recalibration on ordinary slices may improve marginal coverage under the relevant calibration assumptions, but it does not guarantee coverage for atypical perturbations in regions that the measurements do not constrain, which is where the blind set lives: inclusions in $R$ moved $q$ by up to five prior standard deviations. This is the conditional-shift failure of prior-dominated quantities known in the linear setting \citep{knapik2011,hadji2021}; the score's use is to say in advance which perturbations a deployed system cannot vouch for.

\paragraph{Limitations.} The pre-specified score models detection through the total residual only; one score serves both learned systems; the physics-based baseline uses the decoder's model-error covariance and the Laplace spread. Seventeen pre-specified criteria per system were evaluated without multiplicity correction, and the pre-specified intervals resample errors only. Images are $64\times64$ and measurements are simulated, with Gaussian rather than Poisson noise under attenuation. There is one data source, one VAE (latent dimension 32, one training run, warm-started Laplace fits) and one scalar decision quantity, so failures of shape, topology or small features are untested. Errors are inclusions, voids and transplants at chosen locations; attenuation was studied at three strengths. Variation is resampled across noise draws and base slices but not across physical objects: the interleaved split places every sample tube in training and testing, so object-level resampling is not possible with these data. The objects are real CT reference reconstructions, but the restricted-access measurements were simulated from them; no measurement was acquired under the access conditions studied. The conclusions refer to the two inference systems, the slices of the 2DeteCT sample tubes, the error classes and the simulated access conditions studied; validation on independently acquired objects and real restricted-access measurements is needed before any wider claim. Natural next steps are a second decision quantity, recorded sinograms with Poisson noise, independent objects, richer priors, system-specific predictors for ensemble methods, the physics-based baseline as a standard comparator, and measurement design aimed at shrinking the blind set.

\paragraph{Data and code availability.} 2DeteCT is available under CC BY 4.0 \citep{kiss2023}. Code, frozen predictions, realised trial records and the dated planning documents (the analysis plan, the robustness amendment and the seven post-hoc analysis plans, with their dates, hashes and every logged deviation) are deposited on Zenodo at \href{https://doi.org/10.5281/zenodo.23392686}{doi:10.5281/zenodo.23392686} (all versions; version~1.9 adds the seventh post-hoc plan and its results).

\section*{Acknowledgements}
The author conceived the study, defined its research questions and scope, and directed all stages of the work. An AI assistant, Claude (Anthropic; model Claude Opus~5.5), was used under the author's direction. It wrote the analysis code, ran the computations, produced the figures and tables, and drafted the analysis plans and the text of the manuscript. The author reviewed and approved each analysis plan before its computation, reviewed the results and their interpretation, and directed the revisions of the manuscript. The analysis code is deposited with the data (see Data and code availability), so that every computation can be inspected and rerun. The author takes full responsibility for the content.

\bibliographystyle{abbrvnat}
\bibliography{refs}

@article{kiss2023,
  author = {Kiss, Maximilian B. and Coban, Sophia B. and Batenburg, K. Joost and van Leeuwen, Tristan and Lucka, Felix},
  title = {{2DeteCT} -- A large {2D} expandable, trainable, experimental computed tomography dataset for machine learning},
  journal = {Scientific Data}, volume = {10}, pages = {576}, year = {2023}, doi = {10.1038/s41597-023-02484-6}}

@article{knapik2011,
  author = {Knapik, B. T. and van der Vaart, A. W. and van Zanten, J. H.},
  title = {Bayesian inverse problems with {G}aussian priors},
  journal = {The Annals of Statistics}, volume = {39}, number = {5}, pages = {2626--2657}, year = {2011}, doi = {10.1214/11-AOS920}}

@article{hadji2021,
  author = {Hadji, Amine and Szab{\'o}, Botond},
  title = {Can we trust {B}ayesian uncertainty quantification from {G}aussian process priors with squared exponential covariance kernel?},
  journal = {SIAM/ASA Journal on Uncertainty Quantification}, volume = {9}, number = {1}, pages = {185--230}, year = {2021}, doi = {10.1137/19M1253010}}

@article{gugushvili2018,
  author = {Gugushvili, S. and van der Vaart, A. W. and Yan, D.},
  title = {Bayesian inverse problems with partial observations},
  journal = {Transactions of A. Razmadze Mathematical Institute}, volume = {172}, number = {3, Part A}, pages = {388--403}, year = {2018}, doi = {10.1016/j.trmi.2018.09.002}}

@book{hansen2008,
  author = {Hansen, Lars Peter and Sargent, Thomas J.}, title = {Robustness}, publisher = {Princeton University Press}, year = {2008}}

@article{armstrong2021,
  author = {Armstrong, Timothy B. and Koles{\'a}r, Michal},
  title = {Sensitivity analysis using approximate moment condition models},
  journal = {Quantitative Economics}, volume = {12}, number = {1}, pages = {77--108}, year = {2021}, doi = {10.3982/QE1609}}

@article{evans2006,
  author = {Evans, Michael and Moshonov, Hadas}, title = {Checking for prior-data conflict},
  journal = {Bayesian Analysis}, volume = {1}, number = {4}, pages = {893--914}, year = {2006}, doi = {10.1214/06-BA129}}

@article{bhadra2021,
  author = {Bhadra, Sayantan and Kelkar, Varun A. and Brooks, Frank J. and Anastasio, Mark A.},
  title = {On hallucinations in tomographic image reconstruction},
  journal = {IEEE Transactions on Medical Imaging}, volume = {40}, number = {11}, pages = {3249--3260}, year = {2021}, doi = {10.1109/TMI.2021.3077857}}

@article{angelopoulos2023,
  author = {Angelopoulos, Anastasios N. and Bates, Stephen}, title = {Conformal prediction: A gentle introduction},
  journal = {Foundations and Trends in Machine Learning}, volume = {16}, number = {4}, pages = {494--591}, year = {2023}, doi = {10.1561/2200000101}}

@inproceedings{ovadia2019,
  author = {Ovadia, Yaniv and Fertig, Emily and Ren, Jie and Nado, Zachary and Sculley, D. and Nowozin, Sebastian and Dillon, Joshua V. and Lakshminarayanan, Balaji and Snoek, Jasper},
  title = {Can you trust your model's uncertainty? {E}valuating predictive uncertainty under dataset shift},
  booktitle = {Advances in Neural Information Processing Systems}, volume = {32}, year = {2019}}

@inproceedings{kingma2014,
  author = {Kingma, Diederik P. and Welling, Max}, title = {Auto-encoding variational {B}ayes},
  booktitle = {International Conference on Learning Representations (ICLR)}, year = {2014}, note = {arXiv:1312.6114}}

@inproceedings{bora2017,
  author = {Bora, Ashish and Jalal, Ajil and Price, Eric and Dimakis, Alexandros G.},
  title = {Compressed sensing using generative models},
  booktitle = {Proceedings of the 34th International Conference on Machine Learning}, series = {PMLR}, volume = {70}, pages = {537--546}, year = {2017}}

@article{iglesias2013,
  author = {Iglesias, Marco A. and Law, Kody J. H. and Stuart, Andrew M.}, title = {Ensemble {K}alman methods for inverse problems},
  journal = {Inverse Problems}, volume = {29}, number = {4}, pages = {045001}, year = {2013}, doi = {10.1088/0266-5611/29/4/045001}}

@article{emerick2013,
  author = {Emerick, Alexandre A. and Reynolds, Albert C.}, title = {Ensemble smoother with multiple data assimilation},
  journal = {Computers \& Geosciences}, volume = {55}, pages = {3--15}, year = {2013}, doi = {10.1016/j.cageo.2012.03.011}}

@article{browning2026,
  author = {Browning, Alexander P. and Flegg, Jennifer A. and Murphy, Ryan J.}, title = {A cautionary tale of model misspecification and identifiability},
  journal = {Bulletin of Mathematical Biology}, volume = {88}, pages = {5}, year = {2026}, doi = {10.1007/s11538-025-01573-4}}

@misc{gianni2026,
  author = {Gianni, Mario}, title = {Detectability and coverage under restricted access: linear-{G}aussian pilots of the blind set},
  howpublished = {Zenodo, Version 1.0 [technical note, code and data]}, year = {2026}, note = {\doi{10.5281/zenodo.23190701}}, doi = {10.5281/zenodo.23190701}}

\appendix
\section{Evidence protocol}\label{app:protocol}

\paragraph{Pre-specification.} Predictions, decision criteria, thresholds and outcome interpretations were fixed in a dated analysis plan; the prediction file was hashed (SHA-256) and the plan finalised by the author before any realised trial. The plans, their dates and the hashes are recorded in a dated project log and in freeze logs written by the pipeline, but not with an external registry; we therefore describe the study as \emph{pre-specified and committed by hash before trials}. Deviations affecting the main study were a crash fix in the control run and a restart with single-threaded libraries (both before any result was inspected), and an erratum found later: the plan described the model-error covariance as ``rank 60 plus diagonal'', whereas the frozen code (whose hash was recorded) used the 16-parameter spectrum. These are pre-specified decision criteria, not formal tests of the null hypothesis $\rho<0.5$. Before freezing, each learned system had to show at most 15\% false alarms and at least 85\% coverage of $q$ on the test slices without errors.

\begin{figure}[t]\centering
\includegraphics[width=\linewidth]{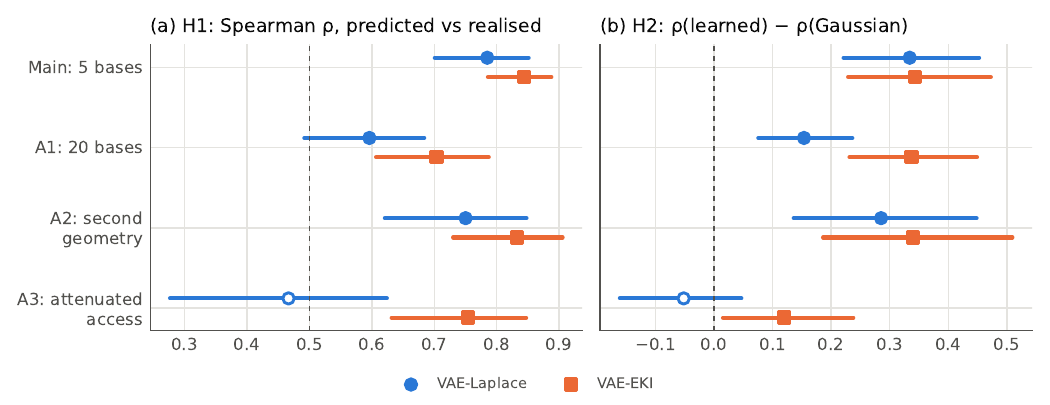}
\caption{H1 and H2 across the main study (Main: bounded reach, 5 base slices) and the robustness study (A1: same set-up, 20 base slices; A2: second geometry; A3: attenuated access). Filled markers: pre-specified criterion met; open: not met. Bars: 95\% bootstrap intervals resampling errors, which understate uncertainty across base slices. Dashed lines: decision thresholds.}\label{fig:forest}
\end{figure}

\paragraph{Robustness study.} Specified in a dated amendment finalised before any of its trials. New criteria: \textbf{H4}, within reach, $\rho$(score, excess) $\geq0.3$ and greater than the Gaussian prediction's; \textbf{H5}, within $R$, the learned score has smaller mean absolute error than the Gaussian prediction against the raw blind rate; \textbf{H6} (attenuation), excess blindness decreases from the deepest ring to the intermediate ring and from the intermediate to the outermost. A1 reused the main-study predictions. A2 recalibrated from scratch and passed the pre-freeze check (false alarms 0.08, 0.10, 0.065; coverage 0.885, 0.88, 0.98) before freezing new predictions. For A3 the attenuation strength was fixed from the model alone as the smallest value on a $\times4$ grid at which the Gaussian posterior standard deviation of central $q$ kept at least half its prior value ($F=128{,}000$); none of the pre-specified grid values $\{500,2000,8000\}$ satisfied the rule, and the grid was extended with the rule unchanged, before any A3 computation on data (a logged deviation). A3 passed the pre-freeze check (false alarms 0.065, 0.06, 0.055; coverage 0.885, 0.86, 0.88). VAE-EKI met all 14 robustness criteria; VAE-Laplace missed 4 (Table~\ref{tab:add}, Fig.~\ref{fig:forest}). With 20 slices, Laplace's within-reach scores were close to the observed rates (mean scores 0.03--0.06; raw rates 0.01--0.03); within $R$ its scores exceeded the observed rate in A1 (0.49, 0.67) by about as much as the Gaussian scores (0.36) fell below it, which is why H5 was not met there. Under attenuation the scores for the deep centre exceeded Laplace's observed blind rate there (0.58 against 0.38).

\begin{table}[t]\centering\small
\caption{Robustness study: pre-specified decision criteria (excess blind rate; 95\% bootstrap intervals resampling errors).}\label{tab:add}
\begin{tabular}{llcc}\toprule
Study & Hypothesis & VAE-Laplace & VAE-EKI\\\midrule
A1 & H1 $\rho\geq0.5$ & 0.596 [0.492, 0.684] \pass & 0.704 [0.607, 0.787] \pass\\
(20 slices) & H2 $\Delta\rho>0$ & $+$0.154 [0.076, 0.236] \pass & $+$0.337 [0.231, 0.449] \pass\\
 & H4 within reach $\rho\geq0.3$ & 0.290 [0.167, 0.410] \fail & 0.608 [0.486, 0.713] \pass\\
 & H4 within reach $\Delta\rho>0$ & $+$0.354 [0.197, 0.510] \pass & $+$0.746 [0.566, 0.924] \pass\\
 & H5 $\Delta$MAE in $R<0$ & $-$0.009 [$-$0.048, 0.030] \fail & $-$0.121 [$-$0.153, $-$0.089] \pass\\\midrule
A2 & H1 & 0.750 [0.621, 0.848] \pass & 0.833 [0.730, 0.905] \pass\\
(second & H2 & $+$0.285 [0.136, 0.448] \pass & $+$0.340 [0.186, 0.510] \pass\\
geometry) & H4 $\rho\geq0.3$ & 0.641 [0.472, 0.771] \pass & 0.747 [0.590, 0.850] \pass\\
 & H4 $\Delta\rho>0$ & $+$0.559 [0.292, 0.815] \pass & $+$0.667 [0.407, 0.925] \pass\\
 & H5 & $-$0.062 [$-$0.095, $-$0.029] \pass & $-$0.130 [$-$0.176, $-$0.085] \pass\\\midrule
A3 & H1 & 0.466 [0.277, 0.624] \fail & 0.754 [0.632, 0.847] \pass\\
(attenuated & H2 & $-$0.052 [$-$0.161, 0.047] \fail & $+$0.120 [0.015, 0.238] \pass\\
access) & H6 ring 0 $>$ ring 1 & $+$0.200 [0.104, 0.304] \pass & $+$0.331 [0.200, 0.469] \pass\\
 & H6 ring 1 $>$ ring 3 & $+$0.062 [0.025, 0.100] \pass & $+$0.042 [0.006, 0.079] \pass\\\bottomrule
\end{tabular}
\end{table}

\paragraph{Post-hoc analysis plans.} After the main and robustness outcomes were known, seven analysis plans were written, each approved and hash-committed before its own computation and each with outcome descriptions fixed in advance: (1) simpler predictors, the three bootstraps and the attenuation-strength sensitivity; (2) the physics-based comparison at all strengths; (3) the score's approximations, calibration, adjusted rankings, a noise-only physics baseline and sensitivity to $\kappa$ and to the linearisation points; (4) the components of the score and their dependence in the realised systems; (5) the trial-level analysis, the linear-Gaussian reference, the location control, the prior--population check, $\kappa=1$ and the flagging use; (6) the detection law at the rank of the null covariance, leave-one-slice-out estimates, the agreement of the learned and physics-based scores, flag budgets from 5\% to 50\%, and the physics-based baseline with $\kappa=1$; (7) refitted bootstrap intervals for the cross-validated log-score gains, their sensitivity to the fold partition, and cross-validation on held-out base slices. One addition after the fifth plan's results were known, the v1 $-$ physics difference under bounded reach (computed with the unchanged second-plan code), is logged as a deviation. As a check on ties, Kendall's $\tau_b$ ordered the learned and Gaussian predictions as Spearman's $\rho$ did in every study and system.

\paragraph{Failed configurations.} Three configurations failed the pre-freeze calibration check before the one reported. (1) \emph{Contiguous splits.} With slices 1--800 for training, 801--900 for calibration and 901--1000 for testing, calibration did not transfer (false alarms 51--68\%, coverage 61--74\% on the test block): the hidden region was denser in that block (mean $q$ 0.151 against about 0.11). (2) \emph{High-rank model-error covariance.} A rank-60 empirical covariance of the decoder's data-space residuals over-fitted the detector null (false alarms on test 68--75\%); the 16-parameter spectrum transferred. (3) \emph{Too few slices.} With 1,000 slices, false alarms on test were 15--20\%; with 2,000, 3--3.5\%.

\section{Further results}\label{app:more}

\begin{table}[t]\centering\footnotesize
\caption{What the predictor is and is not (third post-hoc analysis plan; all errors). CITL: calibration-in-the-large, mean score minus observed raw blind rate. $\Delta$Brier: Brier score of the learned prediction (v1) minus that of the Gaussian prediction (negative favours v1). Partial $\rho$: rank correlation with realised blindness after adjusting for the unreached-region indicator and effect size $|c^\top\delta|$ (location within the reached region is not adjusted for). $\rho_0$: physics baseline with no VAE-derived input. Brackets: two-way bootstrap 95\% intervals.}\label{tab:D}
\setlength{\tabcolsep}{3pt}\resizebox{\linewidth}{!}{\begin{tabular}{llcccccc}\toprule
Study & System & CITL v1 & $\Delta$Brier (v1 $-$ Gauss) & partial $\rho$: v1 & Gauss & physics & $\rho$: physics / $\rho_0$\\\midrule
Main & Laplace & $-$0.08 & $-$0.008 [$-$0.033, 0.018] & $+$0.58 [0.01, 0.69] & $+$0.04 & $+$0.62 & 0.80 / 0.66\\
 & EKI & $-$0.09 & $-$0.016 [$-$0.035, 0.004] & $+$0.66 [0.00, 0.74] & $+$0.07 & $+$0.73 & 0.87 / 0.69\\
\midrule
A1 & Laplace & $+$0.04 & $-$0.005 [$-$0.015, 0.005] & $+$0.26 [0.05, 0.33] & $+$0.09 & $+$0.25 & 0.59 / 0.61\\
 & EKI & $+$0.02 & $-$0.013 [$-$0.024, $-$0.003] & $+$0.51 [$-$0.02, 0.59] & $-$0.01 & $+$0.58 & 0.74 / 0.55\\
\midrule
A2 & Laplace & $-$0.01 & $-$0.005 [$-$0.019, 0.008] & $+$0.58 [0.18, 0.69] & $+$0.13 & $+$0.65 & 0.77 / 0.59\\
 & EKI & $-$0.03 & $-$0.013 [$-$0.027, $-$0.002] & $+$0.67 [0.12, 0.77] & $+$0.09 & $+$0.66 & 0.81 / 0.66\\
\midrule
A3, $F{=}128$k & Laplace & $-$0.02 & $+$0.025 [0.005, 0.052] & $+$0.04 [$-$0.33, 0.43] & $+$0.19 & $+$0.47 & 0.68 / 0.69\\
 & EKI & $-$0.03 & $+$0.004 [$-$0.015, 0.026] & $+$0.56 [0.05, 0.66] & $+$0.36 & $-$0.13 & 0.44 / 0.47\\
\midrule
A3, $F{=}32$k & Laplace & $+$0.02 & $+$0.039 [$-$0.006, 0.086] & $+$0.38 [0.06, 0.56] & $+$0.47 & $-$0.08 & 0.33 / 0.17\\
 & EKI & 0.00 & $+$0.028 [$-$0.017, 0.077] & $+$0.52 [0.21, 0.63] & $+$0.61 & $-$0.04 & 0.39 / 0.32\\
\midrule
A3, $F{=}512$k & Laplace & $+$0.13 & $+$0.037 [0.018, 0.060] & $+$0.12 [0.02, 0.24] & $+$0.15 & $-$0.02 & 0.54 / 0.54\\
 & EKI & $+$0.09 & $+$0.019 [0.003, 0.040] & $+$0.07 [0.02, 0.14] & $+$0.09 & $+$0.01 & 0.66 / 0.66\\
\bottomrule\end{tabular}}\end{table}

\begin{table}[t]\centering\footnotesize
\caption{Partial $\rho$ within reach with a location control (fifth post-hoc plan): errors within reach, adjusted for the rank of $|c^\top\delta|$ and a natural cubic spline in the distance to the unreached disc (A3: to the tube centre). $\Delta$: v1 $-$ physics. Brackets: two-way bootstrap 95\% intervals; $^*$: interval excludes zero. n.d.: not defined. $^\dagger$n.i.: not interpretable (almost all within-reach outcomes are zero; the estimate lies at an interval endpoint).}\label{tab:adjdist}
\setlength{\tabcolsep}{4pt}\begin{tabular}{llccc}\toprule
Study & System & v1 & physics & $\Delta$\\\midrule
Main & Laplace & $+$0.69$^*$ [0.019, 0.776] & $+$0.69 [$-$0.066, 0.798] & 0.00 [$-$0.104, 0.221]\\
 & EKI & $+$0.75$^*$ [0.651, 0.814] & $+$0.78$^*$ [0.718, 0.832] & $-$0.04 [$-$0.103, 0.016]\\
A1 & Laplace & $+$0.34$^*$ [0.076, 0.410] & $+$0.30 [$-$0.143, 0.478] & $+$0.04 [$-$0.265, 0.435]\\
 & EKI & $+$0.64 [$-$0.063, 0.727] & $+$0.71 [$-$0.089, 0.782] & $-$0.06 [$-$0.180, 0.177]\\
A2 & Laplace & $+$0.64$^*$ [0.230, 0.753] & $+$0.69 [$-$0.106, 0.866] & $-$0.05 [$-$0.348, 0.418]\\
 & EKI & $+$0.75$^*$ [0.134, 0.828] & $+$0.70 [$-$0.224, 0.888] & $+$0.05 [$-$0.261, 0.479]\\
A3, $F{=}128$k & Laplace & $-$0.07 [$-$0.432, 0.389] & $+$0.45 [$-$0.235, 0.736] & $-$0.52 [$-$1.106, 0.557]\\
 & EKI & $+$0.53$^*$ [0.008, 0.662] & $-$0.20 [$-$0.636, 0.434] & $+$0.73 [$-$0.319, 1.144]\\
A3, $F{=}32$k & Laplace & $+$0.42$^*$ [0.091, 0.554] & $-$0.07 [$-$0.491, 0.432] & $+$0.49 [$-$0.073, 0.817]\\
 & EKI & $+$0.44$^*$ [0.130, 0.598] & $-$0.14 [$-$0.556, 0.305] & $+$0.58 [$-$0.002, 0.959]\\
A3, $F{=}512$k & Laplace & n.i.$^\dagger$ & n.i.$^\dagger$ & n.i.$^\dagger$\\
 & EKI & n.d. & n.d. & n.d.\\\bottomrule
\end{tabular}
\end{table}

\begin{table}[t]\centering\footnotesize
\caption{Refitted cross-validation (seventh post-hoc plan; log-score gains in nats per trial). H1: gain from adding v1 to the model with physics, plan-5 folds (by error, base-slice intercepts), refitted two-way intervals. H2: range of that gain over 101 random fold partitions. H3: errors and base slices held out together (5 error folds $\times$ 5 slice groups formed by CT block; in Main each group is one slice), common intercept; gains of physics and v1 over the structural model, and of v1 beyond physics, with refitted two-way intervals. M0: mean log score of the structural model under H3 / under the plan-5 design. Held-out slices come from the same sample tubes as the training slices. In the main study, with five base slices, the H3 intervals are very wide and the estimates unstable; they should not be read as evidence about performance on new images. $^\ddagger$n.i.: not interpretable (refitted intervals at $F=512{,}000$; see Table~\ref{tab:trial}).}\label{tab:H}
\setlength{\tabcolsep}{2.5pt}\resizebox{\linewidth}{!}{\begin{tabular}{llcccccc}\toprule
 & & \multicolumn{2}{c}{v1 beyond physics, plan-5 folds} & \multicolumn{4}{c}{held-out base slices and errors (H3)}\\\cmidrule(lr){3-4}\cmidrule(lr){5-8}
Study & System & H1 (refitted) & H2 range & physics over structure & v1 over structure & v1 beyond physics & M0: H3 / plan 5\\\midrule
Main & Laplace & $-$0.002 [$-$0.025, 0.016] & $-$0.005 to 0.000 & $+$0.175 [$-$0.306, 3.098] & $+$0.176 [$-$0.424, 3.152] & $-$0.020 [$-$1.103, 0.025] & $-$0.94 / $-$0.23\\
 & EKI & $-$0.001 [$-$0.071, 0.010] & $-$0.003 to 0.000 & $+$1.191 [$-$0.320, 4.163] & $+$1.197 [$-$0.520, 3.994] & $-$0.194 [$-$2.401, 0.010] & $-$2.66 / $-$0.13\\
A1 & Laplace & $-$0.002 [$-$0.022, 0.005] & $-$0.002 to 0.000 & $+$0.055 [0.007, 0.156] & $+$0.056 [0.008, 0.155] & $-$0.001 [$-$0.020, 0.003] & $-$0.20 / $-$0.17\\
 & EKI & 0.000 [$-$0.019, 0.002] & $-$0.001 to 0.000 & $+$0.297 [$-$0.004, 0.939] & $+$0.298 [$-$0.003, 0.938] & 0.000 [$-$0.092, 0.012] & $-$0.67 / $-$0.12\\
A2 & Laplace & $-$0.001 [$-$0.019, 0.013] & $-$0.001 to 0.000 & $+$0.058 [0.022, 0.278] & $+$0.058 [0.021, 0.283] & $-$0.003 [$-$0.069, 0.007] & $-$0.38 / $-$0.25\\
 & EKI & 0.000 [$-$0.024, 0.013] & $-$0.002 to $+$0.001 & $+$0.053 [0.014, 0.320] & $+$0.053 [0.014, 0.326] & $-$0.003 [$-$0.081, 0.008] & $-$0.37 / $-$0.22\\
A3, $F{=}128$k & Laplace & $+$0.004 [$-$0.007, 0.026] & $+$0.003 to $+$0.006 & $+$0.008 [$-$0.005, 0.044] & $+$0.010 [$-$0.009, 0.055] & $-$0.004 [$-$0.052, 0.017] & $-$0.53 / $-$0.35\\
 & EKI & $+$0.024 [$-$0.002, 0.067] & $+$0.020 to $+$0.028 & $+$0.022 [$-$0.004, 0.091] & $+$0.032 [$-$0.012, 0.103] & $+$0.011 [$-$0.068, 0.056] & $-$0.50 / $-$0.32\\
A3, $F{=}32$k & Laplace & $+$0.020 [$-$0.016, 0.047] & $+$0.013 to $+$0.027 & $-$0.008 [$-$0.132, 0.016] & $+$0.008 [$-$0.081, 0.064] & $-$0.003 [$-$0.099, 0.180] & $-$0.55 / $-$0.27\\
 & EKI & $+$0.035 [$-$0.002, 0.096] & $+$0.027 to $+$0.040 & $-$0.007 [$-$0.061, 0.021] & $+$0.011 [$-$0.105, 0.076] & $-$0.035 [$-$0.260, 0.096] & $-$0.59 / $-$0.33\\
A3, $F{=}512$k & Laplace & n.i.$^\ddagger$ & $-$0.050 to 0.000 & 0.000 [$-$0.099, 0.018] & $+$0.001 [$-$0.173, 0.018] & $-$0.002 [$-$0.128, 0.035] & $-$0.10 / $-$0.06\\
 & EKI & n.i.$^\ddagger$ & $-$0.057 to $+$0.003 & $+$0.001 [$-$0.195, 0.028] & $+$0.006 [$-$0.195, 0.025] & $-$0.006 [$-$0.149, 0.026] & $-$0.11 / $-$0.02\\\bottomrule
\end{tabular}}
\end{table}

\begin{table}[t]\centering\footnotesize
\caption{Components of the score (fourth post-hoc analysis plan; all errors; raw rates). Detection: Spearman $\rho$ of the analytic detection component with the realised alarm rate, and calibration-in-the-large (CITL, mean component value minus realised rate; average level only) of the analytic and Monte-Carlo (MC) components. Non-coverage: CITL of the analytic component (calibrated posterior spread) and of the MC component (sampling variability of the linearised estimator). Swaps (a sensitivity analysis, not a causal decomposition): $G=\rho(A)-\rho(M)$, the gap in ranking realised blindness between the analytic and MC products, with two-way bootstrap 95\% interval; $\rho(A)-\rho(H_{\det})$ and $\rho(A)-\rho(H_{\mathrm{cov}})$, the loss from swapping in the MC detection or coverage component. Joint vs product: mean over errors of the realised blind rate minus the product of realised non-coverage and non-detection rates, and Spearman $\rho$ between the two. Error-only, two-way and block intervals for every entry are in the deposited analysis files.}\label{tab:E}
\setlength{\tabcolsep}{3pt}\resizebox{\linewidth}{!}{\begin{tabular}{llccccccccccc}\toprule
 & & \multicolumn{3}{c}{detection} & \multicolumn{2}{c}{non-coverage CITL} & \multicolumn{3}{c}{swaps} & \multicolumn{2}{c}{joint vs product}\\
\cmidrule(lr){3-5}\cmidrule(lr){6-7}\cmidrule(lr){8-10}\cmidrule(lr){11-12}
Study & System & $\rho$ & CITL & CITL MC & posterior & sampling & $G$ [2-way] & $A{-}H_{\det}$ & $A{-}H_{\mathrm{cov}}$ & mean diff. & $\rho$\\\midrule
Main & Laplace & 0.83 & $-$0.03 & $+$0.37 & $-$0.11 & $-$0.16 & $+$0.16 [$-$0.17, 0.26] & $-$0.01 & $+$0.35 & $+$0.003 & 0.951\\
 & EKI & 0.83 & $-$0.03 & $+$0.37 & $-$0.12 & $-$0.17 & $+$0.18 [$-$0.13, 0.26] & $-$0.02 & $+$0.38 & $+$0.003 & 0.937\\
A1 & Laplace & 0.94 & $-$0.08 & $+$0.33 & $+$0.02 & $-$0.02 & $-$0.01 [$-$0.14, 0.05] & $+$0.01 & $+$0.09 & $-$0.002 & 0.941\\
 & EKI & 0.94 & $-$0.08 & $+$0.32 & $+$0.02 & $-$0.03 & $+$0.12 [$-$0.15, 0.19] & $-$0.03 & $+$0.28 & 0.000 & 0.931\\
A2 & Laplace & 0.93 & $-$0.07 & $+$0.38 & $-$0.06 & $-$0.11 & $+$0.14 [$-$0.10, 0.23] & $-$0.01 & $+$0.31 & $-$0.002 & 0.939\\
 & EKI & 0.94 & $-$0.12 & $+$0.33 & $-$0.11 & $-$0.16 & $+$0.15 [$-$0.16, 0.23] & $+$0.05 & $+$0.34 & $-$0.021 & 0.949\\
A3, $F{=}128$k & Laplace & 0.81 & $+$0.08 & $+$0.52 & $-$0.01 & $-$0.02 & $-$0.21 [$-$0.42, 0.13] & $-$0.21 & $+$0.01 & $-$0.006 & 0.956\\
 & EKI & 0.72 & $+$0.10 & $+$0.54 & $-$0.02 & $-$0.03 & $+$0.18 [$-$0.11, 0.36] & $+$0.19 & $-$0.01 & $-$0.002 & 0.965\\
A3, $F{=}32$k & Laplace & 0.82 & $+$0.08 & $+$0.53 & $+$0.05 & $+$0.08 & $+$0.38 [0.07, 0.57] & $+$0.33 & $+$0.06 & $+$0.007 & 0.922\\
 & EKI & 0.72 & $+$0.02 & $+$0.47 & $+$0.01 & $+$0.04 & $+$0.37 [0.07, 0.60] & $+$0.33 & $+$0.01 & $+$0.009 & 0.969\\
A3, $F{=}512$k & Laplace & 0.64 & $-$0.03 & $+$0.36 & $+$0.13 & $+$0.12 & $+$0.05 [0.00, 0.13] & $+$0.05 & 0.00 & $-$0.003 & 1.000\\
 & EKI & 0.54 & $-$0.02 & $+$0.37 & $+$0.09 & $+$0.07 & 0.00 [0.00, 0.01] & 0.00 & 0.00 & $-$0.005 & 0.981\\\bottomrule
\end{tabular}}
\end{table}

\paragraph{Components (fourth plan).} Table~\ref{tab:E} gives the component analysis. The swap analysis replaces the analytic detection or non-coverage component by its Monte Carlo counterpart; products of per-error means ranked realised blindness like the original score ($\rho$ within 0.01 of v1). Under bounded reach the Monte Carlo non-coverage probability within reach is below 0.001 for 87--92\% of errors (a check on the predictions alone), non-zero mainly for errors with a large effect on $q$, which are also the most detectable; within reach, the score built with it ranked realised blindness negatively ($\rho=-0.05$ to $-0.22$).

\paragraph{Dependence in the realised systems.} We computed a Mantel--Haenszel odds ratio of non-detection given non-coverage at the trial level, stratified by error: $\sum ad/n\big/\sum bc/n$, with $a$, $b$, $c$, $d$ the counts of an error's trials that are undetected and uncovered, undetected and covered, detected and uncovered, and detected and covered, and $n$ its number of trials. No continuity correction was applied and stratum-specific ratios were never formed, so zero cells create no undefined stratum estimates; the pooled ratio is undefined when $\sum bc=0$, such bootstrap replicates were dropped, and an interval is reported only when more than 50 of the 5,000 replicates were defined. The ratios were imprecise and had no consistent direction (0 to 2.6); their two-way intervals included 1 except for EKI at $F=512{,}000$, where the ratio was 0 in every replicate, a sparse-cell effect.

\begin{table}[h]\centering\scriptsize
\caption{Post-hoc differences in $\rho$ with all three 95\% intervals (error-only; two-way, errors $\times$ base slices; block, errors $\times$ 10-slice CT blocks). 5,000 replicates. Studies: Main, main study (bounded reach, 5 base slices); A1, same set-up with 20 base slices; A2, second geometry; A3, attenuated access at strength $F$. Predictors: v1, pre-specified; v2, full-diagnostic. Rows for $F=32{,}000$ and $512{,}000$ are from the second post-hoc analysis plan; within reach at $F=512{,}000$ the EKI outcome is constant, so its $\rho$ is not defined (n.d.).}\label{tab:appB}
\setlength{\tabcolsep}{3pt}
\resizebox{\linewidth}{!}{\begin{tabular}{lll cccc ccc}\toprule
 & & & \multicolumn{4}{c}{v2 $-$ physics} & \multicolumn{3}{c}{v2 $-$ v1}\\\cmidrule(lr){4-7}\cmidrule(lr){8-10}
Study & System & Errors & est. & error-only & two-way & block & est. & two-way & block\\\midrule
Main & Laplace & all & 0.002 & [$-$0.011, 0.011] & [$-$0.015, 0.019] & [$-$0.015, 0.020] & 0.020 & [$-$0.051, 0.064] & [$-$0.052, 0.064]\\
 & Laplace & within reach & 0.000 & [$-$0.029, 0.020] & [$-$0.036, 0.059] & [$-$0.035, 0.060] & 0.044 & [$-$0.173, 0.134] & [$-$0.171, 0.134]\\
 & EKI & all & 0.004 & [$-$0.001, 0.009] & [0.000, 0.009] & [0.000, 0.009] & 0.031 & [0.001, 0.063] & [0.001, 0.063]\\
 & EKI & within reach & 0.009 & [$-$0.002, 0.021] & [$-$0.001, 0.019] & [$-$0.001, 0.020] & 0.069 & [0.000, 0.134] & [0.000, 0.132]\\
\midrule
A1 & Laplace & all & $-$0.028 & [$-$0.057, $-$0.001] & [$-$0.062, 0.020] & [$-$0.064, 0.019] & $-$0.036 & [$-$0.144, 0.067] & [$-$0.146, 0.074]\\
 & Laplace & within reach & $-$0.069 & [$-$0.136, $-$0.006] & [$-$0.150, 0.052] & [$-$0.149, 0.051] & $-$0.085 & [$-$0.392, 0.167] & [$-$0.386, 0.179]\\
 & EKI & all & $-$0.002 & [$-$0.016, 0.010] & [$-$0.014, 0.009] & [$-$0.014, 0.009] & 0.033 & [$-$0.020, 0.073] & [$-$0.021, 0.071]\\
 & EKI & within reach & $-$0.005 & [$-$0.040, 0.025] & [$-$0.035, 0.022] & [$-$0.035, 0.022] & 0.078 & [$-$0.162, 0.167] & [$-$0.158, 0.166]\\
\midrule
A2 & Laplace & all & 0.009 & [$-$0.009, 0.031] & [$-$0.034, 0.058] & [$-$0.033, 0.038] & 0.030 & [$-$0.165, 0.181] & [$-$0.156, 0.153]\\
 & Laplace & within reach & 0.019 & [$-$0.021, 0.067] & [$-$0.076, 0.131] & [$-$0.068, 0.079] & 0.063 & [$-$0.384, 0.367] & [$-$0.382, 0.322]\\
 & EKI & all & 0.000 & [$-$0.022, 0.018] & [$-$0.026, 0.024] & [$-$0.024, 0.018] & $-$0.025 & [$-$0.186, 0.123] & [$-$0.184, 0.152]\\
 & EKI & within reach & 0.000 & [$-$0.046, 0.041] & [$-$0.058, 0.056] & [$-$0.056, 0.040] & $-$0.055 & [$-$0.505, 0.283] & [$-$0.539, 0.337]\\
\midrule
A3 ($F{=}128$k) & Laplace & all & $-$0.095 & [$-$0.223, 0.017] & [$-$0.340, 0.211] & [$-$0.364, 0.238] & 0.114 & [$-$0.018, 0.194] & [$-$0.036, 0.207]\\
 & Laplace & within reach & $-$0.219 & [$-$0.453, 0.005] & [$-$0.672, 0.396] & [$-$0.698, 0.495] & 0.246 & [$-$0.030, 0.397] & [$-$0.077, 0.409]\\
 & EKI & all & 0.269 & [0.130, 0.425] & [$-$0.042, 0.409] & [$-$0.061, 0.415] & $-$0.047 & [$-$0.172, 0.110] & [$-$0.173, 0.116]\\
 & EKI & within reach & 0.550 & [0.267, 0.816] & [$-$0.094, 0.803] & [$-$0.156, 0.796] & $-$0.098 & [$-$0.377, 0.255] & [$-$0.371, 0.270]\\
\midrule
A3 ($F{=}32$k) & Laplace & all & 0.182 & [0.082, 0.305] & [0.026, 0.311] & [0.026, 0.313] & $-$0.072 & [$-$0.186, 0.076] & [$-$0.185, 0.072]\\
 & Laplace & within reach & 0.339 & [0.144, 0.561] & [0.041, 0.574] & [0.035, 0.566] & $-$0.120 & [$-$0.351, 0.171] & [$-$0.346, 0.170]\\
 & EKI & all & 0.218 & [0.105, 0.351] & [0.048, 0.356] & [0.052, 0.361] & $-$0.078 & [$-$0.204, 0.069] & [$-$0.203, 0.076]\\
 & EKI & within reach & 0.396 & [0.182, 0.621] & [0.084, 0.641] & [0.089, 0.632] & $-$0.128 & [$-$0.386, 0.150] & [$-$0.373, 0.174]\\
\midrule
A3 ($F{=}512$k) & Laplace & all & 0.041 & [$-$0.001, 0.141] & [$-$0.001, 0.130] & [$-$0.001, 0.131] & $-$0.001 & [$-$0.003, 0.000] & [$-$0.003, 0.000]\\
 & Laplace & within reach & 0.269 & [0.000, 0.490] & [0.000, 0.472] & [0.000, 0.479] & 0.000 & [0.000, 0.000] & [0.000, 0.000]\\
 & EKI & all & $-$0.001 & [$-$0.002, 0.000] & [$-$0.002, 0.000] & [$-$0.002, 0.000] & 0.000 & [$-$0.001, 0.000] & [$-$0.001, 0.000]\\
 & EKI & within reach & n.d. & & & & n.d. & & \\
\bottomrule\end{tabular}}\end{table}

\end{document}